\documentclass[11pt]{article}

\usepackage[final]{acl}

\usepackage{times}
\usepackage{latexsym}

\usepackage[T1]{fontenc}

\usepackage[utf8]{inputenc}

\usepackage{microtype}

\usepackage{inconsolata}

\usepackage{graphicx}

\usepackage{amsmath}
\usepackage{booktabs}
\usepackage{multirow}
\usepackage{makecell}
\usepackage{enumitem}
\newcommand{\std}[1]{\,$\pm$\,#1}

\title{From Production Traffic to Post-Training: Building a Self-Hosted LLM That Covers the Corporate Request Mix}

\author{
    \textbf{Olga Tsymboi}, \textbf{Dmitrii Stoianov}, \textbf{Ramil Latypov}, \textbf{Danil Taranets},  
    \textbf{Daniil Dryabin}, \\ 
    \textbf{Mikhail Gashkov}, \textbf{Viktor Zelenkovskiy}, \textbf{Aleksandr Fida}, \textbf{Gleb Alektorov}, \textbf{Nikita Gulyakov} \\
    \textbf{Arthur Babkin}, \textbf{Aleksandr Medvedev}, \textbf{Pavel Gein}, \textbf{Anatolii Potapov} \\
    T-Tech \\
 \small{
   \textbf{Correspondence:} \href{mailto:anatolii.s.potapov@gmail.com}{anatolii.s.potapov@gmail.com}
 }
}

\begin{document}
\maketitle
\begin{abstract}
Data-residency constraints force enterprises to self-host LLMs, but
continuous adoption of newer models without decommissioning their
predecessors expands the serving fleet, fragmenting a finite GPU
pool. We consolidate traffic from over 200 internal applications onto
a single model by closing quality gaps identified through
production error analysis along three axes: instruction following,
function-calling, and internal task distribution.
Quality is tracked by offline benchmarks stratified to production
traffic and scored by deterministic verifiers or calibrated LLM judges.
Rather than optimising all objectives jointly, which introduces
cross-domain reward interference, we train a separate GRPO expert per
axis and merge them via two-stage SLERP. Each expert's reward exposes a
distinct failure mode, namely semantic collapse, over-calling, and
verbosity hacking, each requiring a domain-specific fix. In
non-reasoning mode the recipe surpasses a ${\sim}7\times$ larger by
total parameters baseline on the in-house Arena with 69.6 to 65.8, instruction following with 0.85 to 0.83,
and function-calling with 0.79 to 0.77, while lifting general dialogue benchmarks. 
The
model absorbs 50\% of platform traffic, 116M requests per month, at a
fraction of the serving cost.
\end{abstract}

\section{Introduction}

Recent open-weight LLMs have closed the quality gap with proprietary
systems on typical enterprise
workloads~\citep{qwen3technicalreport,dubey2024llama,deepseekai2024},   
and under data-residency and regulatory constraints locally deployed
models are often the only
option~\citep{wu2023bloomberggpt,galbandara2025}. In a large enterprise,
local deployment means one finite GPU pool shared by hundreds of
applications. Each team picks the model that fits its task; switching
later creates friction, new generations arrive quarterly while old
models cannot be retired, and the effective price per token rises as
the zoo fragments.

We consolidate traffic onto a single model by closing the gaps that
prevent migration via dedicated post-training. The adapted model absorbs 50\% of platform traffic,
116M requests per month from over 200 internal applications, six months
after rollout and is cheap enough to update every production cycle.
Because the long tail of applications is created and retired faster
than any one accumulates enough traffic for reliable A/B testing, the
model must retain broad capabilities for unseen workflows, making
evaluation two-sided: diagnose and repair live weaknesses while
preserving generalization.

Error analysis of production traffic surfaces three improvement axes.
Instruction-following and formatting errors account for 37.9\% of
failures (Appendix~\ref{sec:error-analysis}), making strict constraint satisfaction the largest single
lever. Function calling is critical: tool-equipped requests account for
${\sim}$12\% of traffic. A distributional gap remains between
high-frequency task types and long-tail workflows on the one hand and
public training distributions on the other. We address all three in a
single post-training stage without regressing general capabilities and
make three contributions.

This paper presents a methodology for building internal benchmarks from
production traffic, scored by deterministic verifiers or calibrated LLM
judges and validated against human annotators.

A modular post-training recipe trains a separate RL expert per weak
axis and combines them by weight-space merging. Three reward-hacking
failure modes that preclude joint multi-objective training are
documented alongside the domain-specific fix each requires.

An open-weight checkpoint trained with the same recipe but without
internal data is released\footnote{\url{https://huggingface.co/t-tech/T-pro-it-2.1}}, on public benchmarks it scores close to the
deployed version, confirming that the recipe, not proprietary data,
drives the gains.

\section{Related Work}
Russian LLM development has mostly followed two directions: training Russian
  models from scratch~\citep{kuratov2019russian,zmitrovich2023family} and
  adapting multilingual backbones~\citep{tikhomirov2024facilitating,
  nikolich2024vikhr,tpro2,mamedov2025giga}. These efforts improved
  general Russian generation, but reproducible post-training recipes
  targeting agentic skills have not been published for this language.
  Our focus is not fluency, but reliable instruction execution,
  structured tool use, and stable multi-step behaviour.

  \paragraph{Instruction following (IF) and function calling (FC).}
  Both are often treated as language-independent, but this breaks down
  in Russian. IF is harder to check: morphology, case, casing,
  punctuation, and free word order mean one English-style rule has many valid Russian
  forms, so English IF verifiers from AutoIF~\citep{dong2024autoif},
  IFBench~\citep{pyatkin2025generalizing}, and VerIF~\citep{peng2025verif}
  help as templates but cannot be reused as is. FC adds a schema
  constraint: prompts can be localized, but function names, parameter
  keys, and enumeration values are part of the API and must stay fixed.
  Recent multilingual tool-use work~\citep{chen2024enhancing,luo2026lost}
  shows the failure is mostly here: errors come from execution-interface
  violations, not from misunderstanding intent, with parameter-value
  language mismatch the main failure mode, and inference-time fixes do not
  recover English-level performance. We therefore handle both axes at the
  data level, not at inference time: we generate data natively in Russian
  via schema-aware synthetic pipelines~\citep{liu2024apigen,liu2024toolace,
  prabhakar2025apigenmt,xu2025toucan} with the Tool-N1
  reward~\citep{zhang2025tooln1}.

  \paragraph{Post-training recipe.}
  Modern post-training pipelines like Tulu~3~\citep{lambert2024tulu3} and
  DeepSeek-R1~\citep{guo2025deepseekr1} build on RLHF~\citep{ziegler2019finetuning,ouyang2022instructgpt}
  and use RLVR with GRPO and its
  variants~\citep{schulman2017ppo,shao2024deepseekmath,yu2025dapo,zheng2025gspo}. In our
  setting the reward signal is the bottleneck. IF, FC, and general chat
  each pull the model in a different direction, and each reward has a
  trivial shortcut. The same failure modes occur in multi-objective and
  rubric-based RL~\citep{ichihara2025mogrpo,gunjal2025rar,
  huang2025rubricanchors,liu2025openrubrics,shen2026rrd,
  zhang2025chasingtail}, so joint training is fragile. We instead train
  one expert per skill and merge in weight space~\citep{wortsman2022modelsoups,ilharco2023taskarithmetic}, a step now standard in large post-training systems~\citep{dubey2024llama,cohere2024commanda,
  ahmadian2024ayaexpanse}.

  \paragraph{Internal evaluation.}
  Open benchmarks such as Arena-Hard~\citep{li-etal-2025-arena-hard} and
  WildBench~\citep{wildbench2024} provide reproducible scoring of
  open-ended chat. We extend this line with a stratified internal
  benchmark matched to production traffic.

\section{In-house traffic and Arena}\label{sec:main_arena}

We build an internal benchmark from production LLM-platform traffic
and score it with an automated Arena pipeline.
The benchmark is constructed in two stages:
first, a sampling stage that selects a diverse yet representative
subset from ${\sim}100$k monthly queries, and
second, a judging stage that routes each query through a task
classifier and applies a task-specific evaluation scheme.
The methodology is not limited to a one-month window, we use this
interval because it aligns with our production cycle, but the
pipeline applies to any period.
We summarize two key takeaways below.

\paragraph{How to get diversity without drifting from production?}

We quantify diversity as the mean pairwise TF-IDF cosine distance and representativeness as the
Jensen-Shannon distance from the production pool along four
dimensions: queried model, prompt length, service, and task taxonomy. The two objectives are in tension because
in-house traffic is dominated by templated requests, where variable
fragments are substituted into a small set of prompt templates.

Standard approaches fail in opposite directions, as shown in
Table~\ref{tab:sampling-comparison}. Uniform random sampling preserves
the production distribution but inherits near-duplicate structure, while diversity-first methods: greedy max-min and {\#}InsTag~\citep{lu-etal-2024-instag} pursue diversity at the expense of representativeness, distorting service distribution by treating templated
near-duplicates as genuine diversity.

Our template-aware sampler masks variable tokens, groups
near-identical normalized prompts via LSH~\citep{broder1997minhash}, selects within each
template by greedy max-min over variable spans, and allocates
budget as $\sqrt{\text{count}}$. It attains the highest prompt diversity,
$0.953$, while keeping JS distances
substantially lower than pure diversity sampling, the only method that improves coverage without
sacrificing representativeness.

\begin{table}[t]
\centering
\footnotesize
\setlength{\tabcolsep}{3pt}
\begin{tabular}{lrrrrr}
\toprule
\textbf{Method} & \textbf{Dist.} & \textbf{JS\textsubscript{mdl}} & \textbf{JS\textsubscript{len}} & \textbf{JS\textsubscript{svc}} & \textbf{JS\textsubscript{tax}}\\
\midrule
random         & 0.653 & \textbf{0.001} & \textbf{0.083} 
               & \textbf{0.001} & \textbf{0.000} \\
greedy max-min & \underline{0.944} & 0.403 & 0.295 & 0.683 & 0.019 \\
{\#}InsTag     & 0.874 & 0.342 & 0.181 & 0.479 & 0.145 \\
template-based & \textbf{0.953} & \underline{0.122} & \underline{0.098} 
               & \underline{0.281} & \underline{0.015} \\
\bottomrule
\end{tabular}
\caption{Sampling methods compared by average pairwise distance,
Dist., higher is more diverse, and JS distance from the production
pool for queried model JS\textsubscript{mdl}, prompt length
JS\textsubscript{len}, service JS\textsubscript{svc}, and task type JS\textsubscript{tax}, lower
is closer to production.}
\label{tab:sampling-comparison}
\end{table}

\paragraph{Does one judging recipe fit all tasks?}

The answer is no. We score the benchmark with Arena-Hard-Auto
\cite{li-etal-2025-arena-hard} using DeepSeek-V3-0324
\cite{deepseekai2024} as judge. A
uniform side-by-side (SBS) judge agreed poorly with expert
annotators, Cohen's $\kappa=0.62$, and the disagreement tracks
task type. On objective tasks such as classification and
information extraction, the judge and the human should share a reference answer, on open-ended tasks such as summarization and
content generation, ``good'' is underspecified and the two raters
weigh different criteria.

We therefore route each request through an LLM task classifier,
matching human consensus in $90.6$\,--\,$99.6\%$ of cases (see Table~\ref{tab:taxonomy-stability}), and judge each segment with its own
scheme. Classification and information extraction, roughly
$63.2\%$ of traffic, are scored reference-based against gold
answers generated by Kimi-K2.5
\cite{kimiteam2025kimik2openagentic} and verified by annotators,
$97.7\%$ and $85.2\%$ accepted unmodified. Open-ended tasks 
keep pairwise SBS judging but augment it with per-example
RubricHub-style checklists \cite{rubrichub2025} split into
objective and subjective criteria, Appendix~\ref{sec:app_inh_arena}. The best recipe is task-dependent,
Table~\ref{tab:judge-human-agreement}: summarization is judged
best with the checklist as contextual guidance for a single
SBS verdict, $\kappa=0.68$, whereas
content generation benefits from explicit per-criterion grading
with an overall verdict, $\kappa=0.79$.

Overall, the final task-specific pipeline substantially outperforms the uniform-SBS baseline, lifting $\kappa$ from 0.63 to 0.88 on reference-based and from 0.57 to 0.72 on open-ended content-generation subtasks (Table ~\ref{tab:judge-initial-vs-final}).

\begin{table}[!htbp]
\centering
\footnotesize
\setlength{\tabcolsep}{4pt}
\begin{tabular}{lccc}
\toprule
\textbf{Eval setup} & \textbf{Ref.-based} & \textbf{Open-ended} & \textbf{Avg.}\\
\midrule
Uniform SBS      & 0.63 & 0.57 & 0.62\\
Task-specific        & 0.88 & 0.72 & 0.85\\
\bottomrule
\end{tabular}
\caption{Cohen's $\kappa$ by task type for the initial uniform-SBS setup and the final task-specific pipeline. The average is weighted by the number of samples per task.}
\label{tab:judge-initial-vs-final}
\end{table}

\section{Training recipe}
\label{sec:recipe}

Our post-training recipe addresses three axes surfaced by production error analysis: instruction following, function calling, and alignment to the internal task distribution. The model builds on Qwen3-32B \citep{qwen3technicalreport} with an adapted Cyrillic-dense tokenizer~\citep{tpro2}, which proved more efficient than the base one, and operates exclusively in non-reasoning mode due to production latency and generation-cost constraints. Training proceeds in three stages. First, a single combined SFT phase exposes the model to all target domains simultaneously, mixing in-house production, general-domain, instruction-following, and function-calling data. This joint checkpoint then serves as the shared starting point for all subsequent RL branches. We keep SFT shared rather than training and merging per-domain SFT experts: at the SFT stage a naive mixture of all domains retains per-domain quality (Table~\ref{tab:sft_results}), so the extra experts and merge buy nothing, and a single combined stage is the simpler choice.
Rather than optimizing one model against all three reward signals jointly, we fork the SFT checkpoint into three independent GRPO~\citep{shao2024deepseekmath} runs, each trained to convergence on a domain-specific reward. The resulting expert checkpoints are combined into a single deployment model via two-stage sequential SLERP merging~\citep{shoemake1985slerp,goddard2024mergekit}.

\paragraph{General expert}
The general expert retains broad capabilities while aligning to the production task mix. This alignment has no single verifiable reward, and a reward model (RM) adapted to in-house preferences does not improve over the general one (Table~\ref{tab:inhouse_rm_results}), so we address it through the data rather than through a dedicated in-house expert. The data combines general-domain samples from a large open-source Russian instruction corpus~\citep{tpro2}, constituting roughly 80\% of the mixture, with in-house samples drawn from production platform logs making up the remaining 20\%. The in-house portion is sampled using the same strategy as the in-house Arena and decontaminated against benchmark data via Min-Hash. Completions are regenerated by Qwen3-235B-A22B-Instruct-2507, which serves both as a strong teacher and as a way to ensure homogeneous target distribution with the general-domain portion. For RL, we train a general RM on this corpus and run GRPO with two additions: a multiplicative length penalty that compares each response's length to a prompt-specific baseline from Qwen3-235B-A22B-Instruct-2507, and an increased KL coefficient to constrain distribution drift (Appendix~\ref{sec:app_general_expert}).

\paragraph{IF expert} Our in-house IF data appears in the general expert mix, but its diversity may be limited, hurting generalization to unseen constraints in the heterogeneous in-house distribution. Since this corpus was not designed for this capability, a dedicated general IF increment is needed. Training data come from a synthetic pipeline adapted from AutoIF to Russian. Starting from 54 hand-written seed constraints, LLM-based augmentation, consistency-based filtering of generated verifiers and test cases, and back-translation validation expand them to 43K verified constraints. Each constraint is attached to samples from this corpus via both user-level and system-level insertion, yielding 26K training examples with validated completions. During the GRPO we use a VerIF-style verifiable constraint-satisfaction reward. However, pure verifier-based training quickly exposed a reward-hacking failure mode~\citep{skalse2022reward}: the model learned to produce minimal, semantically empty completions that satisfied the formal checks. To fix this, we extend the VerIF-style reward with a prompt-specific reward-model quality correction. Completions that pass the verifier but score below the mean reward-model score over teacher completions on the same prompt receive a penalty. This preserves verifier-reward scalability while preventing semantic collapse (Appendix~\ref{sec:app_if_expert}).

\paragraph{Function-Calling Expert}
Human evaluation of FC platform logs points to two causes of score degradation: production tools are frequently documented in Russian, which is typically out-of-distribution for English-centric models, and the model tends not to select the wrong function but rather fails to fill arguments in Russian. This holds for both Qwen3-32B and Qwen3-235B-A22B-Instruct-2507. Because open FC data and evaluation suites are overwhelmingly English-centric and machine translation of FC data is structurally unsafe, we generate FC training data natively in each language via a synthetic pipeline that builds a tool pool and a multi-turn dialogue pool from scratch, yielding 1.2M English and 300K Russian samples (Appendix~\ref{sec:app_fc_expert}). The dialogue pipeline separates planning from simulation: a planner builds a trajectory with judge feedback, then three agents replay it under asymmetric visibility so training targets reflect realistic tool use rather than leaked references. SFT is a coarse warm-up, with the mixture stratified equally between tool-call and text targets and between English and Russian. For GRPO we use a binary Tool-N1~\citep{zhang2025tooln1} exact-match reward, which admits a dominant exploit: when in doubt, emit a call. We correct this through the data distribution rather than the reward: injecting synthetic irrelevance counters over-calling, while the assistant text-target share is tuned separately for multi-turn accuracy. The resulting mixture is 70\% English and 30\% Russian, with 80\% tool-call and 20\% text targets per language and 10\% synthetic irrelevance within the text share.

\section{Evaluation}
\label{sec:evaluation}
\paragraph{Benchmarks}
Since our primary focus is Russian-language evaluation, we mainly report
Russian-adapted versions of IFEval~\citep{ifeval}, MultiChallenge~\citep{deshpande-etal-2025-multichallenge}, and BFCLv3~\citep{berkeley-function-calling-leaderboard}, described in
Appendix~\ref{sec:app_adapted_benchmarks}, and Arena Hard Ru~\cite{ru_arena_hard}, WildChat Hard Ru~\cite{wildchat-hard-ru}. We additionally report AceBench~\citep{chen2025acebench} and
$\tau^2$-bench~\citep{barres2025tau2bench} to probe tool-calling robustness
across languages.
English results, Arena Hard 2~\cite{li-etal-2025-arena-hard, arenahard2024}
and the original IFEval and MultiChallenge, appear in
Appendix~\ref{sec:app_additional_evals}.

The in-house benchmarks are designed to mirror our production distribution and
comprise four tasks (Appendix~\ref{sec:app_internal_benchmarks}):
the in-house Arena, described in Section~\ref{sec:main_arena};
the in-house IFEval, which pairs deterministic verifiers with an LLM-judge
loose metric;
the in-house BFCL, which evaluates against human-curated references in AST
mode; and SmartSearch, which measures function-calling within a multi-step
ReAct~\citep{yao2023react} loop over internal documents.

For the ablation studies in this section we report a subset,
namely in-house Arena, Arena Hard Ru, ruIFEval, and ru/en BFCLv3, spanning
all capability axes; full results across all benchmarks are reported for the final checkpoints.

\paragraph{Is a combined SFT stage sufficient across domains?}

Before proceeding to alignment, we investigated whether a single shared SFT
model can match the per-domain quality of models trained
exclusively on each domain.
\begin{table}[!htbp]
\centering
\resizebox{\columnwidth}{!}{%
\begin{tabular}{lccccc}
\toprule
\multirow{2}{*}{\textbf{Model}}
  & \multicolumn{2}{c}{\textbf{Arena}}
  & \multirow{2}{*}{\textbf{ruIFEval}}
  & \multicolumn{2}{c}{\textbf{BFCL}} \\
\cmidrule(lr){2-3} \cmidrule(lr){5-6}
  & Ru-Hard & In-House
  & & En & Ru \\
\midrule
Qwen3-32B              & 85.76 & 59.49 & 0.770 & 63.13 & 54.03 \\
\quad + Domain SFT     & \underline{91.96} & \underline{62.18} & \textbf{0.798} & \textbf{69.37} & \textbf{58.69} \\
\quad + Shared SFT     & \textbf{92.45} & \textbf{66.00} & \underline{0.787} & \underline{68.59} & \underline{58.18} \\
\bottomrule
\end{tabular}%
}
\caption{SFT ablation results. Domain SFT denotes a model trained on a single domain: general for the arena columns, instruction following for IFEval, and tool calling for the BFCL columns.}
\label{tab:sft_results}
\end{table}
Table~\ref{tab:sft_results} shows that the shared SFT model closely
matches the domain-specific experts across all three evaluation axes confirming that at the SFT stage a naive mixture of data
from all domains is sufficient to retain per-domain quality. However, as discussed in the following
sections, the same approach of simply combining data does not carry
over to the GRPO stage.

\paragraph{Does the reward model need adaptation to in-house data?}

To investigate this, we trained the reward model on mixtures of general-domain and
in-house preference pairs and compared them against the general-only reward
model within the same GRPO pipeline.
\begin{table}[!htbp]
\centering
\resizebox{\columnwidth}{!}{%
\begin{tabular}{lcccc}
\toprule
\multirow{2}{*}{\textbf{Model}}
  & \multicolumn{2}{c}{\textbf{Ru-Arena-Hard}}
  & \multicolumn{2}{c}{\textbf{In-House Arena}} \\
\cmidrule(lr){2-3} \cmidrule(lr){4-5}
  & Score & Avg.\ Len.
  & Score & Avg.\ Len. \\
\midrule
Qwen3-32B                   & 85.76 & 961  & 59.49 & 208 \\
\quad + Shared SFT           & 92.45 & 1332 & 66.00 & 259 \\
\quad + General GRPO         & \textbf{95.26} & 1261 & \textbf{70.73} & 286 \\
\quad + GRPO w/ In-House RM  & \underline{93.51} & 1261 & \underline{68.58} & 362 \\
\bottomrule
\end{tabular}%
}
\caption{Ablation: effect of adapting the reward model to in-house data.
The in-house-adapted RM does not improve over the general RM.}
\label{tab:inhouse_rm_results}
\end{table}
As shown in Table~\ref{tab:inhouse_rm_results}, adapting the reward model to
in-house data does not yield improvements. Notably, the in-house RM
variant produces substantially longer responses on the in-house evaluation,
averaging 362 tokens compared to 286, which suggests that the adapted model
may have absorbed superficial stylistic biases from the in-house preference
data. These results show that a well-trained general reward model already captures sufficient signal; naively mixing in-house preference pairs into the RM training set can introduce noise or distributional artifacts that slightly degrade performance. We therefore retained the general-domain reward model in our final recipe.

\paragraph{Does single-domain GRPO transfer to other domains?}

When training separate GRPO experts for different capability domains, a practical concern is whether the alignment gains on the target domain
carry over to other domains. To examine cross-domain transfer, we performed GRPO
training on three separate domains, general, instruction following, and
tool calling, each starting from the same shared SFT checkpoint.
\begin{table}[!htbp]
\centering
\setlength{\tabcolsep}{4pt}
\resizebox{\columnwidth}{!}{%
\begin{tabular}{@{}l ccccc@{}}
\toprule
\multirow{2}{*}{\textbf{Model}}
  & \multicolumn{2}{c}{\textbf{Arena}}
  & \multirow{2}{*}{\textbf{ruIFEval}}
  & \multicolumn{2}{c}{\textbf{BFCL}} \\
\cmidrule(lr){2-3} \cmidrule(lr){5-6}
  & Ru-Hard & In-House
  & & En & Ru \\
\midrule
Shared SFT           & 92.45 & 66.00 & 0.787 & 68.59 & 58.18 \\
\quad + General (Gen.)          & \textbf{95.26} & \textbf{70.73} & 0.773 & 68.50 & 55.59 \\
\quad + IF                & 94.88 & 64.34 & \textbf{0.827} & 68.19 & 59.78 \\
\quad + FC                & 92.14 & 62.69 & 0.784 & \textbf{72.25} & \textbf{66.73} \\
\bottomrule
\end{tabular}%
}
\caption{Cross-domain transfer of single-domain GRPO experts, each trained
from the shared SFT checkpoint.}
\label{tab:cross_domain}
\end{table}
Table~\ref{tab:cross_domain} demonstrates that each expert improves on its own domain while the remaining domains see little to no benefit. The
General GRPO expert lifts arena scores to 95.26 and 70.73, but IFEval
and tool-calling metrics remain near the SFT baseline. The IF and FC GRPO experts exhibit the same behavior. This shows that single-domain GRPO alignment transfers poorly
across domains: gains remain confined to the capability covered by the
reward signal, and other domains are largely unaffected. To combine the
improvements from all three domains, the final model was obtained by
merging the three domain-specific experts into a single model.

\begin{table*}[!htbp]
\centering
\small
\setlength{\tabcolsep}{3pt}
\resizebox{\textwidth}{!}{%
\begin{tabular}{@{}lcccccccccccc@{}}
\toprule
\multirow{2}{*}{\textbf{Model}} 
& \multicolumn{2}{c}{\textbf{Arena}} 
& \multicolumn{2}{c}{\textbf{IFEval}} 
& \multicolumn{3}{c}{\textbf{BFCL}} 
& \multirow{2}{*}{\textbf{ACE}} 
& \multirow{2}{*}{$\boldsymbol{\tau^2}$} 
& \multirow{2}{*}{\textbf{SS}} 
& \multirow{2}{*}{\textbf{ruMC}}
& \multirow{2}{*}{\textbf{ruWC}} \\
\cmidrule(lr){2-3} \cmidrule(lr){4-5} \cmidrule(lr){6-8}
& Ru Hard & Inh. & Ru & Inh. & Ru & En & Inh. & & & & & \\
\midrule
T-pro-2.1 internal 
& \underline{93.87} 
& \textbf{69.57} 
& 0.799 
& \textbf{0.85} 
& \underline{65.96} 
& \textbf{72.27} 
& \textbf{0.79} 
& \textbf{73.50} 
& 37.60 
& \underline{0.557} 
& 34.1 
& \underline{80.7} \\

T-pro-2.1 public 
& 93.76 
& \underline{66.8} 
& \textbf{0.807} 
& \underline{0.83} 
& \textbf{66.84} 
& \underline{72.15} 
& \underline{0.78} 
& \underline{72.70} 
& 35.20 
& 0.546 
& \underline{34.8} 
& 78.9 \\

Qwen3-235B-A22B-Instruct-2507 
& \textbf{96.87} 
& 65.83 
& \underline{0.803} 
& \underline{0.83} 
& 64.42 
& 72.13 
& 0.77 
& 70.20 
& \textbf{40.97} 
& \textbf{0.669} 
& \textbf{46.2} 
& \textbf{85.1} \\

T-Pro-2.0 (think) 
& 87.04 
& 61.17 
& 0.687 
& 0.63 
& 50.40 
& 64.38 
& 0.67 
& 63.80 
& 34.97 
& 0.537 
& 31.9 
& 68.5 \\

T-Pro-2.0 (no-think) 
& 90.36 
& 57.46 
& 0.693 
& 0.54 
& 47.47 
& 59.73 
& 0.68 
& 61.20 
& 24.97 
& 0.447 
& 27.8 
& 76.4 \\

Qwen3-32B (think) 
& 87.28 
& 60.46 
& 0.774 
& 0.79 
& 57.33 
& 69.19 
& 0.67 
& 65.00 
& \underline{39.27} 
& 0.491 
& 31.5 
& 59.6 \\

Qwen3-32B (no-think) 
& 85.76 
& 59.49 
& 0.770 
& 0.67 
& 54.03 
& 63.13 
& 0.71 
& 54.60 
& 31.53 
& 0.478 
& 29.3 
& 52.0 \\
\bottomrule
\end{tabular}%
}
\caption{Comparison of models on Dialogue, Instruction Following, and Function Calling benchmarks. Inh.: in-house benchmarks; SS: SmartSearch F1$_{\text{R\&G}}$; ruMC/ruWC: Russian MultiChallenge / WildChat Hard Ru.}
\label{tab:dialogue-if-fc}
\end{table*}
\begin{table*}[!htbp]
\centering
\small
\begin{tabular}{@{}l ccccc@{}}
\toprule
\multirow{2}{*}{\textbf{Merge order}}
  & \multicolumn{2}{c}{\textbf{Arena}}
  & \multirow{2}{*}{\textbf{ruIFEval}}
  & \multicolumn{2}{c}{\textbf{BFCL}} \\
\cmidrule(lr){2-3} \cmidrule(lr){5-6}
  & Ru-Hard & In-House
  & & En & Ru \\
\midrule
(IF + FC) + Gen.   & \textbf{93.37}\std{0.68} & \textbf{68.99}\std{0.77} & \textbf{0.798}\std{0.0032} & \textbf{72.19}\std{0.39} & \textbf{65.73}\std{0.36} \\
(IF + Gen.) + FC   & 91.66\std{0.40} & 65.05\std{0.66} & 0.770\std{0.0036} & 67.83\std{0.44} & 61.25\std{0.26} \\
(FC + Gen.) + IF   & 90.71\std{0.22} & 65.99\std{0.57} & 0.777\std{0.0089} & 70.09\std{0.28} & 64.43\std{0.54} \\
\bottomrule
\end{tabular}%
\caption{Two-stage sequential SLERP merge orderings. Each entry is the mean
$\pm$ standard deviation over five combinations of near-convergence expert
checkpoints merged with identical coefficients.}
\label{tab:merge_order}
\end{table*}

\paragraph{Why merge instead of joint GRPO?}

 An alternative is to optimise all domain rewards in a single joint GRPO
  run. We first study this at 8B, sweeping starting checkpoints, domain
  mixing ratios, and batching schedules
  (Table~\ref{tab:joint_rl_sweep}). The sweep reveals strong cross-domain
  interference: from the mixed-SFT checkpoint, jointly optimising the two
  verifiable-reward domains lifts instruction following (ruIFEval from
  0.731 to 0.801) but pushes function calling below its starting point
  (BFCLv3~EN from 61.2 to 54.5); adding the general reward recovers
  function calling but collapses instruction following below the baseline
  ($0.687$); the only competitive configuration required a general-GRPO
  warm start and a $1.7\times$ budget. 
  We then ran the strongest configuration from each starting checkpoint
at the 32B deployment scale (Table~\ref{tab:joint_vs_merge}), under the same data and evaluation as the
merge recipe. The
8B conclusion transfers: joint GRPO over the three rewards from the
mixed-SFT checkpoint fails to hold all domains simultaneously, with
BFCLv3 falling to 60.88 RU and 70.38 EN against 65.96 and 72.27 for the
merge and ruIFEval to 0.770 against 0.799, while the general-GRPO warm
start reaches parity with the merge only at the $1.7\times$ budget. 
  Joint training thus turns domain balance into a fragile
  hyperparameter search that must be repeated whenever a capability is
  added, whereas the merge recipe extends with one independently trained
  expert and an eval-only coefficient search. We therefore train one
  expert per reward and merge in weight space at the deployment scale
  (Tables~\ref{tab:cross_domain},
  \ref{tab:merge_order}, \ref{tab:joint_vs_merge} and~\ref{tab:merge_operator}).

\begin{table}[!htbp]
\centering
\setlength{\tabcolsep}{4pt}
\resizebox{\columnwidth}{!}{%
\begin{tabular}{@{}l ccccc@{}}
\toprule
\multirow{2}{*}{\textbf{Method}}
  & \multicolumn{2}{c}{\textbf{Arena}}
  & \multirow{2}{*}{\textbf{ruIFEval}}
  & \multicolumn{2}{c}{\textbf{BFCL}} \\
\cmidrule(lr){2-3} \cmidrule(lr){5-6}
  & Ru-Hard & In-House & & En & Ru \\
\midrule
Merge                  & 93.87 & 69.57 & \textbf{0.799} & \textbf{72.27} & \textbf{65.96} \\
Joint, from SFT                    & 94.04 & 70.27 & 0.770 & 70.38 & 60.88 \\
Joint, from Gen.\ GRPO$^{\dagger}$ & \textbf{94.65} & \textbf{70.45} & 0.786 & 72.06 & 65.29 \\
\bottomrule
\end{tabular}%
}
\caption{Joint multi-domain GRPO vs.\ expert merging at the 32B scale.
$^{\dagger}$ trained with a $1.7\times$ larger budget than the other runs.}
\label{tab:joint_vs_merge}
\end{table}

\paragraph{Does merging order matter?}
Because SLERP is non-associative, the order in which experts are composed affects the result.
Merging the two verifiable-reward experts first and then folding in the general expert outperformed the alternative orderings (Table~\ref{tab:merge_order}) by more than the spread we measured across expert checkpoints.
We chose that order empirically, by enumerating all combinations of the three
experts.
Appendix~\ref{sec:app_merging} reports further ablations on the merging operator.

\paragraph{Results and Deployment}

Table ~\ref{tab:dialogue-if-fc} compares the final merged checkpoint with the base Qwen3-32B, T-Pro-2.0, and the ${\sim}7\times$ larger by total parameters Qwen3-235B-A22B-Instruct-2507. Our model is on par with or ahead of the larger model across the in-house Arena, 69.57 compared to 65.83, in-house BFCL, 0.79 to 0.77, ruBFCLv3, 65.96 to 64.42, and AceBench, 73.50 to 70.20, all in non-reasoning mode, at a fraction of the serving cost. On benchmarks where backbone scale dominates, namely ruMultiChallenge which probes long-context memory and self-coherence and SmartSearch which requires open-ended retrieval and synthesis, the recipe substantially narrows the inherited gap: SmartSearch $F_1$ improves from 0.478 to 0.557, surpassing both 32B-scale thinking-mode baselines, and ruWildChat rises from 52.0 to 80.7, within 4.4 points of the ${\sim}7\times$ larger model. The public checkpoint, trained with the same recipe but without the internal-data increment, scores close to the deployed version on all benchmarks except the in-house Arena, and is released as open weights. In production the merged checkpoint serves 116M requests per month, 45 average and 110 peak requests per second, from over 200 internal services on single-GPU FP8 replicas behind vLLM \citep{kwon2023vllm}, 16 to 48 pods, with 95th-percentile latency of 3.2 s and time-to-first-token of 0.3 s. Compared to the ${\sim}7\times$ larger baseline it matches on quality, the deployed model reduces per-token cost by 2.8 to $3.9\times$ on input and output, and up to 4 to $9\times$ for services that previously ran the largest platform models. The few rollbacks came from teams requiring frontier-scale agentic capabilities beyond a 32B dense model.

\section{Conclusion}

We presented a production-driven recipe for consolidating a fragmented
self-hosted LLM fleet into a single deployment model. Traffic analysis across
more than 200 internal applications identified three main gaps: instruction
following, function calling, and alignment to the internal request
distribution. Joint GRPO made these objectives interfere, so we trained one
expert per axis from a shared SFT checkpoint and merged them with two-stage
SLERP.

The main lesson is that enterprise post-training is easier to control when the
objectives are modular. Each axis produced its own failure mode and each required a targeted
fix. This separation makes the recipe easier to debug, audit, and extend.

Together with template-aware traffic sampling and task-specific judging
calibrated against human annotators, the recipe gives a practical path from
production error analysis to deployment. The final Qwen3-32B non-reasoning
model is competitive on target deployment metrics while serving 116M monthly
requests at lower cost. A public checkpoint trained without internal data
shows similar public-benchmark gains, suggesting that the recipe accounts for
much of the improvement.

\section*{Limitations}

\paragraph{Language and deployment scope.}
Our evaluation covers only Russian and English. The in-house benchmarks are
built from Russian-language traffic of a single self-hosted corporate
deployment, while English enters both the training mixtures and the corresponding evaluation
sets. The methodology: traffic-stratified benchmarking, one RL expert per
weak axis, and weight-space merging, is not inherently tied to these
languages or to our organization, though we leave verification on other languages and domains to future work. All quantitative
claims in this paper should therefore be read as validated for Russian and
English only.

\paragraph{Reliance on LLM judges.}
Open-ended quality is scored by LLM judges calibrated against human
annotators on our benchmark distribution.
If the recipe is reused for another deployment, language, or benchmark
distribution, judge quality should be re-calibrated against human annotations. A sensitivity study over four judges is reported in Appendix~\ref{sec:app_additional_evals}.

\paragraph{Deployment and model-family scope.}
The deployment evidence comes from a single organization and a single
self-hosted corporate platform, and all experiments use only the Qwen3 model family. Nothing in the recipe depends on the backbone, but we have not validated it on another family or in another organization.

\section*{Ethical Statement}

\paragraph{Data provenance, privacy, and decontamination.}
All in-house benchmarks are derived from logged traffic of an internal
corporate LLM platform, processed entirely on self-hosted infrastructure
under data-residency constraints; this is also why the system is deployed
locally rather than through a third-party API. Records are deduplicated and
anonymized, and variable spans such as numbers, identifiers, and long
literals are masked during template extraction. No public or external
end-user data is involved, and only internal work-related requests are used.
We explicitly control for contamination between training and evaluation data:
evaluation items are held out from the internal-data increment used to train
the general expert, and exact and template-level duplicates are removed. The
weak-axis experts are trained on synthetic data only, so internal traffic
does not directly supervise them. This separation underpins our main
claim: the improvements from the weak-axis experts and their
weight-space merges are not explained by memorizing the same internal
examples that appear in the benchmarks.

\paragraph{Human annotation.}
Gold answers, benchmark translations, and the task-specific verification
procedures were validated and corrected by professional annotators and domain
experts working as part of this project; their role was data validation and
correction. In the judge-validation studies, each response
pair was independently labeled by three annotators and the consensus taken as
the majority vote. All annotation was performed on internal work-related data
under the privacy and data-handling procedures described above.

\paragraph{Intended use and risks.}
The model is intended for internal enterprise assistance. Like any LLM, it
can produce factually incorrect outputs or violate stated constraints;
deterministic verifiers reduce but do not eliminate this risk. In particular,
tool calls emitted by the model should be gated and validated before
execution in production.
\bibliography{custom}

\appendix
\newpage
\
\newpage

\section{Internal benchmarks}
\label{sec:app_internal_benchmarks}

\subsection{Production error analysis}
\label{sec:error-analysis}

To ground the choice of improvement axes, we sampled $n{=}2{,}500$ LLM-platform traffic
responses and had three annotators assign each its single primary failure to one
of six categories; inter-annotator agreement was Cohen's $\kappa = 0.62$ and the
category label was taken as the majority vote (Table~\ref{tab:error-analysis}).

Two observations motivate the IF axis. Classification is the largest single
category (36.0\%), but instruction-following failures, once formatting (21.0\%)
and non-format (16.9\%) violations are combined, exceed it (37.9\%) and are
additionally checkable by deterministic verifiers, making them a natural target
for a dedicated expert with a verifiable reward (Appendix~\ref{sec:app_if_expert}).

Second, the remaining categories either do not isolate into a single verifiable
axis or are governed by factors a targeted post-training stage cannot move.
Classification is a heterogeneous
downstream symptom rather than a single capability: in the non-reasoning mode
our deployment mandates for latency and cost, a misclassification rarely exposes
a clean rewardable signal, so it does not form a domain expert the way IF and FC
do. Knowledge-base errors track parametric capacity and long-context retention;
they scale with the backbone and are addressed by general-domain training that
reduces hallucination and improves helpfulness, not by a constraint-style
reward: consistent with the residual gap to the ${\sim}7\times$ larger model on
knowledge- and memory-heavy benchmarks (Section~\ref{tab:dialogue-if-fc}).
Language-preference errors are minor for the bulk of production traffic and are
likewise absorbed by the general expert.

This taxonomy covers text-reply failures; function-calling errors are analysed
separately by human evaluation of tool-equipped logs, which form ${\sim}12\%$ of traffic (Table~\ref{tab:taxonomy-stability}),
and motivate the FC axis. That evaluation traces score degradation primarily to Russian-language tool descriptions, which lie out of distribution for English-centric base models, and to argument-filling errors in Russian rather than incorrect function selection, a pattern consistent across both Qwen3-32B and Qwen3-235B-A22B-Instruct-2507.

\begin{table}[t]
\centering
\small
\begin{tabular}{lr}
\toprule
Category & Share \\
\midrule
Classification                    & 36.0\% \\
Formatting (IF)                   & 21.0\% \\
Instruction following, non-format & 16.9\% \\
Knowledge base                    & 19.0\% \\
Language preference               & 5.0\%  \\
Other                             & 2.1\%  \\
\bottomrule
\end{tabular}
\caption{Failure-type distribution over a human-reviewed
sample of $n{=}2{,}500$ in-house traffic responses, one primary
failure per response ($\kappa = 0.62$). Formatting denotes
violations of structural, JSON-schema, or length constraints;
the non-format row covers content-level instruction violations
such as tone, role, enumeration, and lexical prohibitions.}
\label{tab:error-analysis}
\end{table}

\subsection{Inhouse Arena}
\label{sec:app_inh_arena}
\subsubsection{On benchmark data collection}

This section specifies the greedy max-min and template-based sampling procedures summarized in Table~\ref{tab:sampling-comparison}.

\paragraph{Greedy max-min.} Starting from a random seed, the sampler repeatedly adds the query whose minimum cosine distance to the already-selected set is largest, so every new pick is the one most dissimilar from all previous picks. This maximizes the minimum pairwise distance of the selected subset. Since production traffic contains large families of near-identical templated requests, the procedure must place a point in each family to avoid coverage gaps, spending budget on near-duplicates and pulling the sample toward rare queries far from the cluster centres.

\paragraph{Template-based sampling.} A markdown-aware parser splits each prompt into structural segments. Within every segment we mask variable tokens such as numbers, opaque identifiers, and long string literals, which yields a normalized segment sequence. Near-identical sequences are grouped into templates via locality-sensitive hashing, and a majority vote gives one consensus template per group. Greedy max-min then runs within each template over the variable spans only, and the per-template budget scales as $\sqrt{\text{count}}$ so that frequent families neither vanish nor dominate. Unlike fixed-prefix grouping, this merges formatting variants of the same task and separates prompts whose prefixes coincide by chance.

\subsubsection{On benchmark methodology}
\label{app:taxonomy}

\paragraph{Task taxonomy validation.} Each request is assigned to the task taxonomy by an LLM classifier before annotation. On items where annotators reached a task-type consensus, the classifier label matched that consensus in 90.6 to 99.6\% of cases across all four task types, lowest on information extraction, a small fraction of which borders classification. This confirms that the automatic labels reflect how experts categorize the same requests. The taxonomy is also stable over time: re-classifying an independent batch from a non-overlapping window with the same classifier leaves the task-type distribution essentially unchanged, with no category shifting by more than three percentage points (Table~\ref{tab:taxonomy-stability}).
\begin{table}[!htbp]
\centering
\small
\setlength{\tabcolsep}{3pt}
\begin{tabular}{lrrrr}
\toprule
\textbf{Category} & \textbf{Slice~1} & \textbf{Slice~2} & \textbf{$\Delta$~pp} & \textbf{Acc.} \\
\midrule
Classification          & 55.5\% & 52.8\% & $-2.7$ & 99.6\% \\
Summarization           &  8.6\% & 11.4\% & $+2.8$ & 97.8\% \\
Information extraction  &  7.7\% & 10.7\% & $+3.0$ & 90.6\% \\
Content generation      &  5.2\% &  3.7\% & $-1.4$ & 92.1\% \\
\addlinespace
Tool Calling            & 11.8\% & 11.9\% & $+0.1$ & --- \\
Other                   & 11.3\% &  9.5\% & $-1.8$ & --- \\
\bottomrule
\end{tabular}
\caption{Primary task-type distribution on two independent slices of internal requests (LLM classifier labels) and the accuracy (Acc.) of each assigned label against human consensus (on items where annotators reached a consensus). Accuracy covers the four classifier task types only; Tool Calling is detected by a regular expression and Other is the residual category.}
\label{tab:taxonomy-stability}
\end{table}

\paragraph{Task-type preservation under sampling.} We also asked whether each sampling method preserves the task-type composition of production traffic, using the same taxonomy as in Table~\ref{tab:taxonomy-stability}. Table~\ref{tab:sampling-comparison} reports the Jensen--Shannon distance between the benchmark and its reference pool over primary task-type labels, lower is closer to production.

Uniform random and template-based sampling both keep the task-type distribution essentially aligned with production, $0$ and $1.5\%$ respectively. Greedy max-min introduces only a modest taxonomy shift, $1.9\%$, far smaller than its distortion of the service and model dimensions in Table~\ref{tab:sampling-comparison}. {\#}InsTag departs sharply from the reference taxonomy, $14.5\%$: by prioritizing topically diverse queries it over-represents rare task categories, consistent with its elevated JS on the service and model dimensions. Template-based sampling is therefore the only configuration that both maximizes lexical diversity and preserves task-type, length, and service composition close to production.
\paragraph{Different scoring methodologies.}
For open-ended generative tasks (summarization, content generation) we keep pairwise SBS against a baseline, as in Arena-Hard, but augment it with explicit per-example criteria \cite{wildbench2024, zhang2025chasingtail, rubrichub2025} covering instruction following, completeness, factual correctness, and formatting. We generate these checklists with Kimi-K2.5 in a RubricHub-style pipeline \cite{rubrichub2025} and split each item into objective criteria, grounded in the masked request template, and subjective criteria, grounded in the instance-specific variable spans.

To check that the criteria capture what actually drives a verdict, each annotator independently produced both per-criterion verdicts (pass, fail, not applicable) and a separate overall SBS verdict. Criteria were applicable in nearly all cases: annotators marked not applicable in under $2\%$ of summarization and at most $0.4\%$ of content-generation judgments. Reconstructing the overall verdict from per-criterion votes alone recovers almost perfect agreement with the human consensus and clearly beats an objective-only rule. We use an objective-then-subjective Criteria Aggregation (Obj$\rightarrow$Subj) rule: a response that fails any objective criterion loses outright, and the subjective criteria decide only when neither response is eliminated this way; per-annotator labels are then combined by majority over the three annotators. The gap from the full rule to the objective-only variant is largest on content generation, confirming that subjective criteria carry essential signal (Table~\ref{tab:criteria-derived}).

\begin{table}[!htbp]
\centering
\small
\begin{tabular}{llc}
\toprule
\textbf{Task} & \textbf{Aggregator} & \textbf{Cohen's Kappa} \\
\midrule
\multirow{2}{*}{Summ.}
  & Criteria Agg (Obj$\rightarrow$Subj) & \textbf{0.85} \\
  & Objective-only                      & 0.72 \\
\midrule
\multirow{2}{*}{Cont. gen.}
  & Criteria Agg (Obj$\rightarrow$Subj) & \textbf{0.87} \\
  & Objective-only                      & 0.49 \\
\bottomrule
\end{tabular}
\caption{Reconstructing the consensus overall SBS verdict from per-criterion annotator votes alone. The full criteria aggregator recovers almost perfect agreement; removing subjective criteria substantially degrades agreement, especially on content generation.}
\label{tab:criteria-derived}
\end{table}

\subsubsection{Judge validation against human annotations}
\label{sec:judge-validation}

To assess the referee's agreement with the annotators' responses we adopt a naming scheme that makes three design dimensions explicit and keeps them disentangled: the prompt format, the decision source, and the aggregation rule:

\begin{itemize}[leftmargin=1em, itemsep=0pt, topsep=4pt, parsep=0pt]
    \item \textbf{Baseline SBS:} the original pairwise judge without generated criteria.
    \item \textbf{Criteria-Guided SBS:} gives the judge the full checklist as contextual guidance but asks for one SBS verdict.
    \item \textbf{Per-Criterion (Obj$\rightarrow$Subj):} asks the judge to grade model answers against all criteria in a single API call (evaluating each as \textit{Pass / Fail}) and then aggregates those verdicts using the Criteria Agg. (Obj$\rightarrow$Subj) rule.
    \item \textbf{Per-Criterion + Overall Verdict:} asks the judge to grade model answers against all criteria and then provide an arena-style overall verdict from the same structured call.
\end{itemize}

We validate these configurations against expert SBS annotations using DeepSeek-V3-0324 as the judge model.

All metrics in Table~\ref{tab:judge-human-agreement} are means over three independent runs. Following arena-style practice \cite{li-etal-2025-arena-hard}, each response pair is judged twice per run, once in the original A/B order and once reversed, and the two verdicts are combined by numeric averaging, as in the Arena-Hard-Auto positional-bias correction. We report Cohen's Kappa ($\kappa$) rather than raw accuracy because the SBS label distribution is imbalanced; $\kappa$ is our primary metric.

\begin{table}[!htbp]
\centering
\small
\begin{tabular}{llc}
\toprule
\textbf{Task} & \textbf{Judging method} & \textbf{$\kappa$} \\
\midrule
\multirow{4}{*}{Summ.}
  & Baseline SBS                             & 0.61 \\
  & Criteria-Guided SBS                      & \textbf{0.68}  \\
  & Per-Crit. (Obj$\rightarrow$Subj) & 0.53 \\
  & Per-Crit. + Overall & 0.58 \\
\midrule
\multirow{4}{*}{Cont. gen.}
  & Baseline SBS                             & 0.49 \\
  & Criteria-Guided SBS                      & 0.55  \\
  & Per-Crit. (Obj$\rightarrow$Subj) & 0.70 \\
  & Per-Crit. + Overall & \textbf{0.79} \\
\midrule
\multirow{2}{*}{Avg. Open-ended}
  & Baseline SBS                             & 0.57\\
  & Task dependent &  \textbf{0.72} \\
\bottomrule
\end{tabular}
\caption{Judge--human agreement for DeepSeek-V3-0324. The Avg. Open-ended row reports a task-weighted average over the summarization and content generation tasks.}
\label{tab:judge-human-agreement}
\end{table}

Compared to the baseline SBS judge, adding criteria substantially improves agreement: on summarization, Criteria-Guided SBS raises $\kappa$ from $0.61$ to $0.68$; on content generation, Per-Criterion + Overall Verdict raises $\kappa$ from $0.49$ to $0.79$ (Table~\ref{tab:judge-human-agreement}).

As shown in Table~\ref{tab:judge-initial-vs-final}, improvements are observed on both task slices. However, the reference‑based $\kappa$ values correspond to different evaluation targets (SBS preference compared to correctness against gold answers) yet both quantify judge--human agreement.

\subsection{Inhouse IFEval}
\label{sec:app_inh_ifeval}

Production assistants and support-style applications routinely impose explicit constraints on reply form: output format, length budgets, language, allowed and forbidden phrasings, templates, tone, and role-play rules. Public benchmarks like IFEval and IFBench~\citep{pyatkin2025generalizing} probe constraint satisfaction under synthetic, predominantly English prompts unrepresentative of deployed traffic.

\paragraph{Construction.}
The benchmark reuses the AutoIF pipeline~\citep{dong2024autoif} for verifier generation and quality filtering. The only methodological difference is that AutoIF starts from a small set of hand-written seed instructions augmented by an LLM, whereas our prompts come directly from in-house traffic with no seeds — for each sampled prompt, an LLM extracts candidate constraints from scratch, and the AutoIF gates (executable verifier, cross-validation on independent test cases, back-check) are applied unchanged. Approximately 33\% of sampled queries pass the pipeline, yielding a benchmark of 1{,}000 prompts with $\sim$1{,}800 validated verifiers. Because the source prompts are real support-style and structured-output requests, the constraint mix is shifted toward JSON-schema constraints — required key sets, per-field types and value ranges, single-JSON-only output, per-field length and language — alongside the IFEval-style prose categories.

\subsection{Inhouse BFCL}
\label{sec:app_inh_toolcalling}
Existing benchmarks do not match the distribution of a production platform serving diverse internal teams. Tool-equipped requests make up ~12\% (see Table~\ref{tab:taxonomy-stability}) of our traffic, spanning varied domains, toolset sizes, and description verbosity. We built a stratified benchmark matched to this distribution along four axes: domain, toolset size, dialogue stage, and trajectory depth, so that aggregate scores predict real deployed performance.

\paragraph{Construction.}
We follow the BFCL methodology in AST mode~\citep{berkeley-function-calling-leaderboard}, with two adaptations for our production setting. First, tools in our LLM platform traffic are live systems owned by tenant teams; invoking them from a benchmark harness would trigger side effects, and most lack offline substitutes. We therefore evaluate in AST mode only, scoring predicted calls against references structurally and semantically rather than by executed outcome. Second, as in BFCL, references are human-curated: from 1{,}000 sampled examples, expert annotators produce valid (tool, argument, value) configurations and classify each argument as \textit{constrained} (enum, numeric, boolean, or verbatim context string) or \textit{free-text} (free-form natural language). Free-text arguments are scored on presence alone to avoid false negatives from paraphrastic variation.

\paragraph{Coverage filter.}
During curation, roughly 28\% of sampled examples are dropped: cases where annotators could not agree on a stable acceptable set, or whose answer is entirely free-form text unverifiable under AST matching. This retention filter on benchmark ambiguity replaces the implicit filtering that hand-curated \texttt{possible\_answers} provide in BFCL, bounding the scorer's false-positive rate at the cost of reduced coverage on the free-text-only slice of production traffic.

\paragraph{Metric.}
Each candidate call is scored against the curated acceptable set by a deterministic match function that emits per-component verdicts: call-vs-text \textit{decision}, chosen \textit{tool(set)}, \textit{required} arguments presence, \textit{schema} validity, and constrained-argument values. Each component is true, false, or not-scored when inapplicable (e.g.\ no values to check on a text-reply example); the headline \textit{example pass} is the conjunction of all applicable components. No LLM judge is invoked at scoring time.

\paragraph{Limitations of the metric.}
Two limitations merit note. First, free-text argument values are excluded from value matching by construction; a model that hallucinates a plausible but incorrect free-text body is not penalised on this axis. Second, text replies are scored only on the call-vs-text \textit{decision}, not on content. The residual gap to stricter expert-level evaluation including free-text quality is captured by the in-house dialogue arena (Section~\ref{sec:app_inh_arena}).

\subsection{SmartSearch benchmark}
\label{sec:app_smart_search}
Public benchmarks evaluate FC quality in isolation, but a production agent typically issues tool calls inside a longer retrieval-and-reasoning loop. To measure this regime on a realistic in-domain task, we introduce \emph{SmartSearch}, a human-curated benchmark over internal corporate sources.
 
\paragraph{Data collection.} Each evaluation sample is a query and source pair: the query is a natural-language information request authored by a domain expert annotator, and the source is the single internal wiki page that the annotator identified as containing sufficient information to answer it. Every query has exactly one canonical source by construction, which yields an unambiguous ground-truth reference; queries whose evidence was spread across multiple pages were excluded during annotation.
 
\paragraph{Tool space.} The agent is given exactly two tools that wrap an internal smart-search service. The first is a \emph{search} tool that returns a ranked set of text chunks for a query. The second is a \emph{page-QA} tool that takes a free-form question about a specific wiki page and returns an answer; internally, a separate model instance with a clean context receives the full page content together with the question and answers it in a standard QA format, and that answer is returned as the tool result. Restricting the agent to these two tools lets the benchmark compare models on FC ability — query formulation, the choice between broad retrieval and targeted page-level questioning, and grounding of the final answer — rather than on tool-discovery behaviour.
 
\paragraph{Reference construction.} For each sample, a golden answer is produced by prompting Qwen3-235B-A22B-Thinking-2507~\citep{qwen3technicalreport} on the full content of the reference page, and an atomic list of ground-truth claims is then extracted from the golden answer. Both the golden answer and the extracted claim list were reviewed by human annotators and corrected where necessary.

\paragraph{Metrics.} Three metrics are reported. \emph{Recall} measures the fraction of ground-truth claims that an LLM judge marks as present in the agent's answer, with partial matches receiving half credit. The grounding of the answer is captured through two intermediate quantities over the claims the agent produces: \emph{Citation Coverage}, the fraction of those claims that cite some source, and \emph{Citation Correctness}, the fraction of the cited claims whose citation actually supports them. \emph{Grounded Rate} combines the two, measuring the fraction of claims that are both cited and correctly supported. Recall and Grounded Rate are summarised by their harmonic mean, denoted F1$_{\mathrm{R\&G}}$. All LLM-judge protocols were calibrated against independent human annotators: the initial Recall judge (Qwen3-30B-A3B-2507-Thinking with the first version of the prompt) reached $85\%$ agreement with the humans, and after error analysis — prompt revision, switching the judge model to Qwen3-Next-Instruct, and fixing the decoding temperature to $0$ with a repetition penalty of $1.05$ to avoid output loops — agreement rose to $95\%$. The judges used for Citation Coverage and Citation Correctness reached $92\%$ agreement under the same protocol.

\paragraph{Evaluation regime.} The agent operates in a ReAct loop~\citep{yao2023react}, interleaving thought, tool call, and observation steps until it emits a final answer with citations.

\section{Adapted public benchmarks}
\label{sec:app_adapted_benchmarks}

\subsection{Russian IFEval}
To evaluate instruction-following capabilities in Russian, we constructed ruIFEval, a Russian adaptation of the original English IFEval benchmark. We started from the original benchmark samples and manually translated them into Russian with the help of human annotators. During translation, we preserved the intent of each prompt and the structure of the tested instruction-following behavior, while adapting the wording to sound natural in Russian. This adaptation was relatively straightforward because IFEval is largely language-agnostic: most tasks rely on exact, verifiable constraints rather than language-specific morphology, such as case inflection or agreement.

In addition to translating the prompts, we also localized the constraint set used by the benchmark verifiers. These constraints cover different types of instruction-following requirements, such as formatting, keyword inclusion or exclusion, length restrictions, counting constraints, and structural requirements. When a constraint was language-dependent, we adapted it to the Russian setting rather than translating it literally. For example, constraints requiring an exact number of occurrences of a particular character were changed to use Cyrillic analogues, so that the verifier still tests the same type of behavior but in a linguistically appropriate way.

\subsection{Russian BFCLv3}
\label{sec:app_ru_bfcl_v3}
Russian-language FC evaluation is poorly served by existing multilingual resources. AceBench~\citep{chen2025acebench} extends to Chinese but not Russian, and MLCL~\citep{luo2026lost} targets Chinese, Hindi, and Igbo. None offers a Russian split suitable for training-stage decisions in our setting. We therefore localised BFCLv3 into Russian. Unlike at training time, where the scale of the data rules out human verification, evaluation sets are small enough that translation combined with human verification and manual correction is tractable; the resulting set is what we refer to throughout this work as ruBFCLv3. The central concern of the localisation is the preservation of the unambiguity and solvability of each example: a translated user request, target answer, and tool descriptions must not introduce contradictions or new ambiguities. The user request and the associated tools are translated jointly with structured output, automatically generating a translation schema that preserves the original JSON structure of each tool. An ensemble of LLM judges then verifies translation correctness along two axes — absence of logical contradictions and completeness of translation across all semantically meaningful fields. The target answer is translated next, conditioned on the validated request and tools, and is itself verified by LLM judges. After this automatic pipeline, a panel of public models (Qwen3-8B, Qwen3-32B, Qwen3-235B-A22B-Instruct-2507, GLM-4.5, and GLM-4.6; \citealp{glm45}) was scored on the resulting set, and examples on which model disagreement was abnormally high relative to the corresponding English example were flagged for human review and corrected manually. This human-in-the-loop step is what makes translation tractable for an evaluation set but prohibitively expensive at training-data scale.

\subsection{Russian MultiChallenge}
\label{sec:app_ru_multichallenge}

We evaluate models on MultiChallenge~\citep{deshpande-etal-2025-multichallenge}, a multi-turn instruction-following benchmark targeting these long-horizon failures, using both the original English version and a Russian adaptation constructed for this work.

To measure the same capabilities in Russian, we construct a Russian version of MultiChallenge by translating all dialogues, preserving dialogue roles, turn order,
evaluation axis, and binary pass criterion. Rubrics checking a translated surface string are translated; the remaining language-independent rubrics are kept in English so
both versions share identical criteria.

Naive full-dialogue translation failed: models sometimes followed embedded instructions instead of translating, producing omissions and reference shifts. We therefore
translate one turn at a time, providing preceding turns as context. This is especially important for \textsc{Reliable Version Editing}, whose conversations are $2.4\times$
longer on average ($\sim$2{,}300 words, 13 turns).

For all subsets except \textsc{Reliable Version Editing}, we apply an iterative translate--verify--revise pipeline: a separate \texttt{gemini-3.1-pro-preview} \citep{gemini31pro} call checks
each turn against the English source for preserved instructions, facts, references, and rubric conditions; flagged turns are revised for up to five rounds. Expert
annotators then verified every translation. \textsc{Reliable Version Editing} was translated entirely by annotators due to its greater complexity relative to other subsets.

After translating the dialogues, annotators manually reviewed every binary rubric. Most encode a semantic condition invariant to surface language and were kept in
English, our multilingual judge, \texttt{gemini-3.1-pro-preview}, handles English rubrics reliably even when scoring Russian responses. A minority test a surface
form, requiring a verbatim string (e.g.\ It's important to note that'', Yes''/``No'') or forbidding a specific word---and had to be localized to match the Russian
dialogue. In total, 44 of the 273 rubrics were localized---21 in \textsc{Instruction Retention}, 20 in \textsc{Reliable Version Editing}, and 3 in \textsc{Self-Coherence},
with none in \textsc{Inference Memory}, consistent with that axis being purely semantic.

For both languages, models generate responses to the full multi-turn
conversation, and only the final response is evaluated with the
benchmark's instance-level binary rubric. We keep decoding and judging
settings identical across \texttt{en} and \texttt{ru} whenever
possible, so that differences are not confounded by changes in the
generation or evaluation setup.

\section{General Expert}

\label{sec:app_general_expert}
\paragraph{Mitigating response length increase.}

\begin{table}[!htbp]
\centering
\resizebox{\columnwidth}{!}{%
\begin{tabular}{lcccc}
\toprule
\multirow{2}{*}{\textbf{Model}}
  & \multicolumn{2}{c}{\textbf{Ru-Arena-Hard}}
  & \multicolumn{2}{c}{\textbf{In-House Arena}} \\
\cmidrule(lr){2-3} \cmidrule(lr){4-5}
  & Score & Avg.\ Len.
  & Score & Avg.\ Len. \\
\midrule
Qwen3-8B                       & 54.43 & 1116 & 52.23 & 221 \\
\quad + SFT                     & 74.92 & 1388 & 54.81 & 308 \\
\quad + DPO                     & 85.13 & 2064 & 57.44 & 384 \\
\quad + DPO w/ Len. Rebal.    & \underline{85.56} & 2010 & 57.05 & 375 \\
\quad + GRPO w/ Len. Pen.     & 82.66 & 1274 & \textbf{59.09} & 362 \\
\quad + GRPO w/ Len. \& KL    & \textbf{86.02} & 1220 & \underline{58.31} & 320 \\
\bottomrule
\end{tabular}%
}
\caption{Ablation of alignment stages on Qwen3-8B. Score denotes win rate against the baseline; Avg. Len. is the mean response length in tokens. Len. Rebal. refers to rebalancing DPO training  to favor shorter chosen responses. Len. Pen. and KL denote the multiplicative length penalty and increased KL divergence coefficient, respectively}
\label{tab:general_expert_results}
\end{table}
\begin{figure*}[!htbp]
    \centering
    \includegraphics[width=0.7\textwidth]{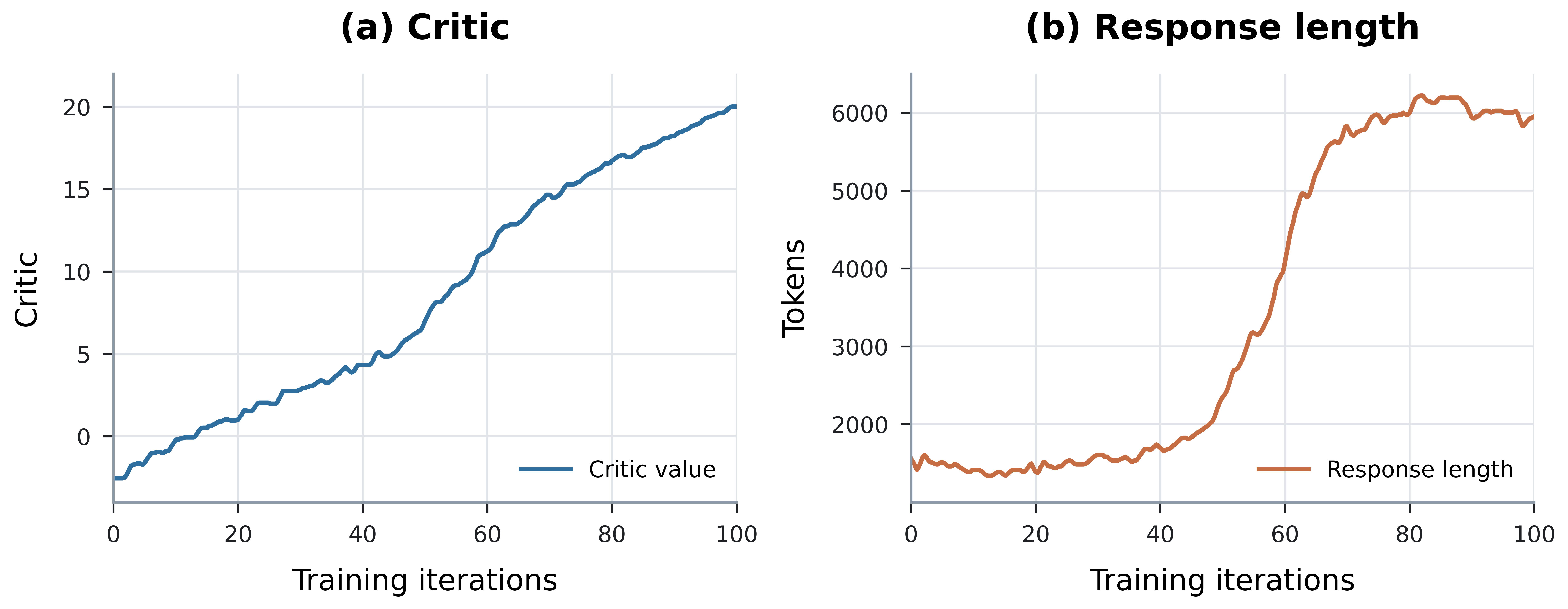}
    \caption{Evolution of GRPO mean reward (left) and mean response length in tokens (right) over training steps. Without length and KL regularization, the model exploits the RM verbosity bias.}
\label{fig:length_explosion}
\end{figure*}
A persistent challenge during alignment was the tendency of trained models to produce increasingly longer responses, as the reward model exhibited a preference for verbose outputs~\citep{singhal2023length,dubois2024lengthcontrolled}. The ablation described below was conducted using Qwen3-8B, once the final recipe was established, it was applied to Qwen3-32B. Our initial DPO~\citep{rafailov2023dpo} runs, while improving quality on arena-based benchmarks, nearly doubled the average response length (from 1388 to 2064 tokens on Ru-arena-hard; see Table~\ref{tab:general_expert_results}), and attempts to rebalance the training set by favoring shorter chosen answers~\citep{park2024disentangling} had negligible effect. Switching to GRPO with only the reward model score as a signal exacerbated the problem: the model learned to hack the length-biased reward~\citep{gao2023scaling} by generating additional self-posed questions and answering them, drifting far from the initial policy's distribution and rendering training ineffective (Figure~\ref{fig:length_explosion}). To address this, we combined two techniques. First, we introduced a multiplicative length penalty that transforms the reward score $R(x,y)$ based on the ratio of the response length $L(y)$ to a baseline length $L_0(x)$ defined as the response length of Qwen3-235B-A22B-Instruct-2507:
$$R(x, y) \mapsto R(x, y)\bigl(1 - \alpha(x, y)\bigr)$$
$$\alpha(x, y) = \operatorname{sgn}\, R(x, y) \cdot \operatorname{clip}\!\left(\frac{\tfrac{L(y)}{L_0(x)} - d_{\min}}{d_{\max} - d_{\min}}\right)$$
where $d_{\min}=0.1$ and $d_{\max}=0.3$ define the bounds of the penalized length deviation range. The $\operatorname{sgn}$ factor ensures correct behavior for both positive and negative reward values, which simpler additive or milder multiplicative variants failed to handle; all alternatives we tested still allowed the model to hack the reward. Second, we increased the KL penalty coefficient from $0.001$ to $0.01$ to constrain distribution drift~\citep{ziegler2019finetuning}; without this, the model stagnated in quality on our benchmarks despite rising reward scores, and also degraded on other experts' domains, undermining the merged model. The combination of both techniques yielded a model that surpassed DPO in benchmark scores, achieving $86.02$ compared to $85.56$ on Ru-arena-hard and $58.31$ compared to $57.05$ on the in-house arena, while producing substantially shorter responses of $1220$ compared to $2010$ and $320$ compared to $375$ tokens, respectively, validating GRPO with joint length and KL regularization as the alignment method of choice.

\section{Instruction-Following Expert}
\label{sec:app_if_expert}
\subsection{Data Creation}

To build specialized IF training data, we used a synthetic data generation pipeline inspired by AutoIF~\cite{dong2024autoif}, adapting it to Russian. This adaptation is necessary because many IF constraints are language-dependent and involve formatting, morphology, punctuation, lexical restrictions, and stylistic requirements. Therefore, all major components of the pipeline were generated and filtered in Russian. The pipeline largely follows the AutoIF data construction procedure, but we instantiate it for the Russian setting and report the concrete filtering thresholds, dataset sizes, and implementation details used in our experiments. We also highlight several practical modifications introduced to improve reproducibility and reduce degenerate or low-quality samples. The pipeline was designed around verifiable instructions whose compliance can be checked automatically by deterministic or semi-deterministic validation functions. The data creation process consisted of several stages.

\paragraph{Constraint generation.}

We started from 54 hand-written seed instruction types from IFEval, AutoIF, and IFBench~\cite{ifeval, dong2024autoif, pyatkin2025generalizing}, translated them into Russian, and covered formatting, lexical, structural, length, and multi-condition constraints. To expand the pool, we repeatedly sampled groups of five constraints as in-context examples and asked an LLM to generate five new verifiable constraints in the same format. This produced about 100k candidates. Exact-match and semantic deduplication reduced them to 72k unique constraints.

\paragraph{Verifier and test-case generation.}

For each constraint, we generated 8 candidate validation functions and 3 synthetic test cases per function. The validators were evaluated against test cases with expected binary labels. We then applied consistency-based filtering. First, we removed test cases for which fewer than half of the validators produced the expected label, since such cases were likely ambiguous, mislabeled, or underspecified. Next, we re-estimated validator reliability on the remaining cases and removed functions that failed to execute or achieved accuracy below 0.5. Finally, we discarded a constraint if fewer than 3 validators or fewer than 5 test cases remained, and additionally required at least 2 positive and 2 negative test cases. After this stage, 50k constraints remained.

\paragraph{Back-translation validation.}

To improve semantic alignment between constraints and validators, we generated a natural-language instruction from each validation function and compared it with the original constraint using cosine similarity. If the similarity was below 0.6, the validator was removed. If more than 60\% of validators for a constraint failed this check, or if all validators were removed, the whole constraint was discarded. This filtered out validators that implemented a different condition from the intended instruction and left 43k constraints.

\paragraph{Mapping constraints to SFT samples.}

Each remaining constraint was randomly attached to 3 SFT samples, producing 131k constraint-augmented samples. This allowed us to combine verifiable IF requirements with diverse underlying tasks instead of training only on synthetic standalone prompts. For each augmented sample, we generated a candidate response and scored it using the prompt-based procedure from AutoIF, retaining only samples with the maximum score of 10. This left 62k samples.

Then we split these samples by constraint placement. For 20k samples, the constraint was inserted into the user instruction. To increase phrasing and positional diversity, we used three strategies: prepending the constraint, appending it, or rewriting the instruction so that the constraint was naturally integrated. The remaining 42k samples were system-level constraint samples, where the constraint was placed in the system instruction.

\paragraph{Completion generation and final selection.}

For each sample, we generated 8 candidate completions and scored them with the corresponding validators. We removed samples for which all completions had validation accuracy below 1, ensuring that each retained sample had at least one fully valid completion. The final dataset contained 10k user-level and 16k system-level constraint samples. For each retained sample, we additionally scored valid completions with our reward model and selected the best one.

\subsection{Training}

We first added IF data during SFT, allowing the model to imitate validated completions in both user-level and system-level settings. However, SFT only provides positive demonstrations and does not directly optimize strict constraint satisfaction under sampling.

As an alignment baseline, we also evaluated DPO on the same IF data. It did not improve IF metrics and increased the average response length by approximately 1.5x, suggesting that preference optimization may encourage verbosity without reliably improving verifiable compliance. This motivated us to move from preference-based alignment to RLVR-style training.

\begin{figure*}[!htbp]
    \centering
    \includegraphics[width=0.7\textwidth]{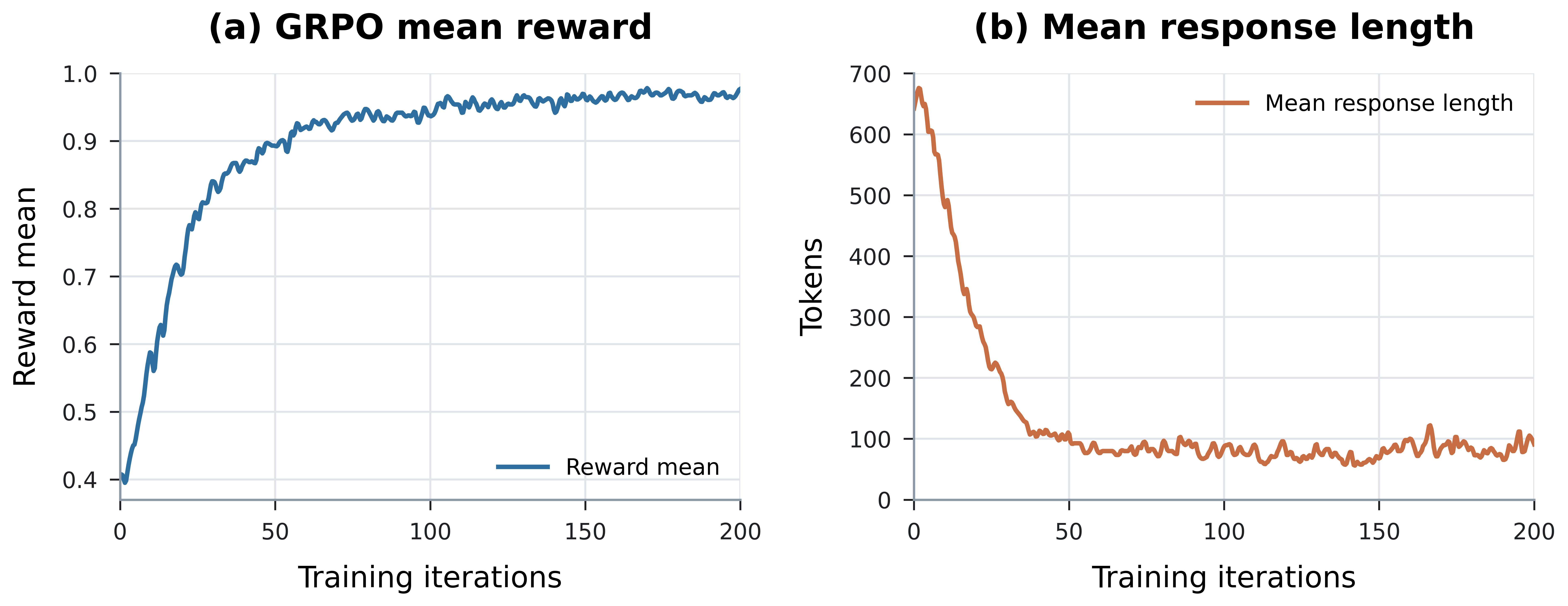}
    \caption{GRPO mean rewards (left) and mean response length in tokens (right) by training steps.}
    \label{fig:reward_hacking_if}
\end{figure*}

We used GRPO-style reinforcement learning with verifiable rewards. Each prompt was associated with validation functions checking whether the generated response satisfied the requested constraints, and the initial reward was based on verifier-derived IF accuracy. This reward led to rapid gains in measured IF accuracy but also exposed reward hacking. The model learned to optimize validators directly instead of producing useful answers. Empirically, the average reward increased while the average response length sharply decreased, as shown in Figure~\ref{fig:reward_hacking_if}. The model discovered shorter and less informative completions that satisfied formal checks but degraded answer usefulness.

A representative example, translated into English, is shown below. The system instruction requires an acrostic whose first letters spell ``EXAMPLE''. Instead of answering the user's question, the model outputs only the required initials. This may satisfy the acrostic validator but is clearly unhelpful because it does not provide the requested answer.

\begin{quote}
\textbf{System:} Write the entire answer as an acrostic: the first letters of the lines must spell the word ``EXAMPLE''.

\textbf{User:} One in five Americans suffers from allergies. In a random sample of 250 Americans, how many people would you expect to suffer from allergies?

\textbf{Assistant:} \\
\begin{tabular}{l}
E \\
X \\
A \\
M \\
P \\
L \\
E \\

\end{tabular}
\end{quote}

This behavior occurs because verification functions typically evaluate only whether a specific constraint is satisfied. They do not necessarily assess whether the response is complete, coherent, helpful, or semantically adequate. To address this issue, we evaluated several combinations of verifiable and semantic reward components, including stronger KL regularization, explicit length penalties, reward-model-based penalties, and an LLM-as-a-Judge penalty based on a 1--10 scoring prompt. Results are presented in Table~\ref{tab:reward_ablation}.

Our final reward follows the IF-RLVR-style formulation~\cite{pyatkin2025generalizing}, extended with a reward-model-based quality correction. Let $V_i$ be the verifiable reward for sample $i$, and $S_i$ be the reward-model score of the generated completion. The final reward $R_i$ is:
\[
R_i =
\begin{cases}
V_i + 1, & \text{if } V_i > 0 \text{ and } S_i > \alpha_i, \\[4pt]
V_i - 0.5, & \text{if } V_i > 0 \text{ and } S_i \leq \alpha_i, \\[4pt]
V_i, & \text{if } V_i \leq 0.
\end{cases}
\]

Here, $\alpha_i$ is a prompt-specific reward-model threshold. Unlike the original formulation, which uses a fixed threshold from the reward-model score distribution, we compare each completion against a prompt-specific quality baseline. We tried experiments based on reference answers (RM / ref.), the current SFT model (RM / SFT), and a stronger teacher model (RM / 235B). The best results were obtained when $\alpha_i$ was set to the average reward-model score of 8 completions from Qwen3-235B-Instruct-2507 on the same prompt. As shown in Table~\ref{tab:reward_ablation}, the reward-model penalty with a prompt-specific teacher baseline gave the strongest and most consistent improvements on ruIFEval. This reward design keeps verifier rewards precise and scalable, while the prompt-specific reward-model threshold penalizes formally valid but low-quality outputs.

\begin{table}[!htbp]
\centering
\footnotesize
\setlength{\tabcolsep}{2.5pt}
\begin{tabular}{lcccc}
\toprule
& \multicolumn{4}{c}{\textbf{ruIFEval}} \\
\cmidrule(lr){2-5}
\textbf{Setup} & \textbf{P-S} & \textbf{P-L} & \textbf{I-S} & \textbf{I-L} \\
\midrule
Qwen3-8B & 0.692 & 0.722 & 0.786 & 0.808 \\
8B SFT & 0.683 & 0.712 & 0.765 & 0.793 \\
\midrule
DPO & 0.659 & 0.763 & 0.750 & 0.790 \\
Len. pen. & 0.711 & 0.737 & 0.794 & 0.817 \\
Judge pen. & 0.673 & 0.718 & 0.764 & 0.797 \\
RM / ref. & 0.707 & 0.738 & 0.791 & 0.814 \\
RM / SFT & 0.712 & 0.748 & 0.793 & \textbf{0.820} \\
RM / 235B & \textbf{0.720} & \textbf{0.750} & \textbf{0.802} & \textbf{0.820} \\
\bottomrule
\end{tabular}
\caption{
Alignment and reward ablation on ruIFEval for 8B model.
P-S/P-L denote strict/loose prompt-level accuracy; I-S/I-L denote strict/loose instruction-level accuracy.
}
\label{tab:reward_ablation}
\end{table}

\section{Function-Calling Expert}
\label{sec:app_fc_expert}
The FC expert targets four capabilities: selecting the correct tool from a heterogeneous pool, populating arguments with appropriate types, chaining dependent calls across turns, and abstaining when no available tool fits. Russian is a particular focus, since open FC training data and evaluation suites are overwhelmingly English-centric. The expert is evaluated on English BFCLv3, AceBench, $\tau^2$-bench and on ruBFCLv3 (Appendix~\ref{sec:app_ru_bfcl_v3}); these score the four capabilities jointly, with the multi-turn subset (Table~\ref{tab:fc_multiturn}) the most direct measure of chaining. BFCLv3 additionally serves as the development benchmark throughout this appendix: all mixture and checkpoint decisions are made on it (ruBFCLv3 enters only the language-split sweep), whereas AceBench and $\tau^2$-bench enter no selection decision, so the per-stage gains on them (Table~\ref{tab:fc_results}) are free of selection effects.

\begin{table}[!htbp]
\centering
\small
\begin{tabular}{lcc}
\toprule
Model & BFCLv3 (en) & BFCLv3 (ru) \\
\midrule
Qwen3-32B (no-think)        & 21.00 & 17.75 \\
~~+ SFT                     & 38.12 & 29.88 \\
~~+ SFT + GRPO              & \textbf{42.88} & \textbf{37.12} \\
\bottomrule
\end{tabular}
\caption{Multi-turn subset accuracy on BFCLv3 (en/ru) across the FC recipe stages. The multi-turn regime, which exercises dependent calls across turns, roughly doubles from base to the SFT-plus-GRPO stage.}
\label{tab:fc_multiturn}
\end{table}

\subsection{Data Creation}

\paragraph{Limitations of existing data.}
FC data is generated synthetically rather than reused from open corpora, for three reasons. First, open FC data is English-only. Second, translating English FC data into Russian is structurally unsafe --- function names, parameter keys, and \texttt{enum} values must remain untouched while user utterances and tool descriptions are rewritten, and the cross-references between an utterance and a selected \texttt{enum} value are routinely severed by translation; \citet{chen2024enhancing} document this for the English--Chinese setting, and \citet{luo2026lost} identify \emph{parameter value language mismatch} as a dominant multilingual FC failure mode. Third, open corpora are short, single-turn, shallow-schema, and narrow in domain. As a preliminary check, SFT was run from the base model on the union of public FC datasets --- xLAM/APIGen~\citep{liu2024apigen}, Hermes~\citep{hermesfcv1}, and ToolACE~\citep{liu2024toolace} --- and evaluated against the base: training on the original English data gave no appreciable gain on English BFCLv3, and a machine-translated Russian version moved ruBFCLv3 by less than a point. Both the tool pool and the dialogue pool are therefore generated from scratch in each target language, following ToolACE, APIGen-MT~\citep{prabhakar2025apigenmt}, and Toucan~\citep{xu2025toucan}.

\paragraph{Tool generation.}
The tool pool is built in two stages. Seed topics are extracted from public FC datasets via BERTopic clustering~\citep{grootendorst2022bertopic} and expanded from 15.5K to 33.5K through self-instruct~\citep{wang2023selfinstruct} with inter-round MinHash deduplication~\citep{broder1997minhash}; a topic is deliberately narrow (e.g.\ ``UTF-8 string operations'') and serves as a diversity knob rather than a taxonomy. Then 15 tools per topic are generated by gpt-oss-120b~\citep{openai2025gptoss} and passed through Merge (semantic deduplication), Refine (description and schema enrichment)~\citep{ye2026feedback}, JSON-Schema validation, and a final MinHash pass; a function name and its required parameters are held invariant throughout, so all derived versions stay call-compatible. Description length is controlled separately: compression passes after Merge and after Refine populate the short end of the range --- Refine in particular inflates descriptions --- giving four length levels per tool. The Russian pool is generated natively rather than translated: the pipeline runs end-to-end in Russian (topics included); the few-shot seed exemplars are translated once by Qwen3-235B-A22B-Instruct-2507 with structured fields and literals held constant, then manually validated --- tractable at exemplar scale, unlike at training-data scale; the Refine prompt is additionally constrained to keep description language consistent, since Russian tools otherwise drift back into English. After deduplication the English run yields 441K tool variants (unique tools $\times$ length levels) and the Russian run 456K; deduplication is applied within each language. 

\begin{figure*}[!htbp]
\centering
\includegraphics[width=0.7\textwidth]{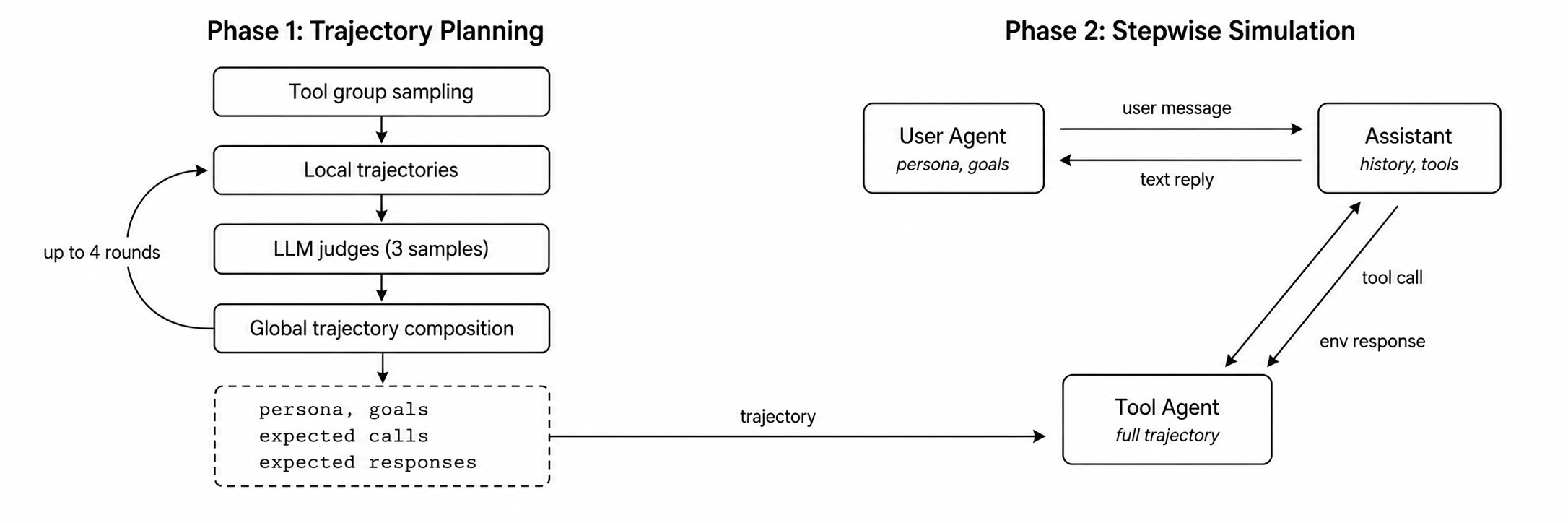}
\caption{Dialogue generation pipeline. A planner constructs a structured trajectory from a sampled tool group and refines it through up to four rounds of judge feedback. The trajectory is then handed to a three-agent stepwise simulation in which the User Agent, Assistant, and Tool Agent operate under asymmetric visibility; flagged erroneous assistant turns remain in context but are never used as training targets.}
\label{fig:dialogue_generation}
\end{figure*}

\paragraph{Dialogue generation.}
Single-pass generation exhibits three failure modes: task degeneration (the user is endowed with information only the tool can provide), mode collapse onto a few templates, and the absence of a reference against which hallucinated arguments can be detected. The pipeline (Figure~\ref{fig:dialogue_generation}) separates planning from execution to mitigate all three, in the spirit of APIGen-MT and ToolACE: conditioning each trajectory on a randomly sampled tool group mitigates mode collapse, fixing the user goal and calls ahead is designed to preclude task degeneration, and the trajectory serves as the reference for checking assistant calls. The planner first builds \emph{local trajectories} per tool group --- persona, goal, expected calls with arguments, and expected environment responses --- each scored by three judge samples (drawn at high temperature to diversify critiques) over up to four revision rounds, then composes them into a \emph{global trajectory} spanning the dialogue with a further such round. In Phase~2, three agents replay it under asymmetric visibility: the User Agent sees only persona and goals, the Assistant only history and available tools, and the Tool Agent --- which alone has the full trajectory --- validates each call and emits only what a real tool would return, never exposing the plan or its ground-truth arguments. User and assistant turns are each sampled multiple times and resolved by judge and majority voting respectively. Turns failing voting or validation are kept in context for later recovery but never used as training targets. All roles are filled by Qwen3-235B-A22B-Instruct-2507, yielding 1.2M English and 300K Russian turn-level training samples, the Russian portion from a native-language run. The tools are not executable, so correctness rests on the planner's reference rather than execution, and a single model family generates and validates the dialogue data (the tool pool comes from a different family, gpt-oss-120b). The evaluation is external to this pipeline --- BFCLv3, AceBench, and $\tau^2$ are public, and ruBFCLv3 is human-verified --- and the deployed merged model ends up ahead of the dialogue generator itself on ruBFCLv3 (65.96 vs.\ 64.42; Table~\ref{tab:dialogue-if-fc}).

\subsection{Training}

The tool and dialogue pools feed two training stages, SFT then GRPO; this subsection describes the recipe and the ablations behind the GRPO mixture.

\paragraph{Model selection.}
The stages are studied in isolation: SFT is run on FC data only from the base model, and GRPO continues from that FC-SFT checkpoint, so each stage's contribution can be attributed cleanly. This differs from the main recipe, where the FC expert branches from the shared multi-domain SFT, so the numbers here are not those of the deployed model. FC is composite, and in preliminary runs the 8B base responded inconsistently to SFT recipe changes, so SFT is studied directly on Qwen3-32B in no-think mode. The GRPO mixture search is run at 8B for cost. Two of its three axes --- the text share and the irrelevance share --- tune a behavioural bias rather than a capability, so unlike the capability gains from SFT their direction is expected to transfer to 32B, where it is re-verified (Table~\ref{tab:grpo_32b_confirm}); the language split is a data-balance choice with no such transfer claim and is simply fixed at 8B.

\paragraph{SFT.}
Each assistant turn is a separate training sample with loss on that turn only. SFT is treated as a coarse warm-up: the mixture is stratified 50/50 between tool-call and free-form text targets --- avoiding both an over-representation of tool calls, which tends to suppress textual fluency, and of text, which dilutes the core skill --- and 50/50 English--Russian, by downsampling the larger English pool. Samples are additionally stratified by total tool-description length: in development runs, controlling for tool count alone still left performance variation across otherwise comparable dialogues, so total tool-description length serves as the difficulty signal, and the mixture is biased toward longer-description samples while retaining short ones for coverage.

\begin{table}[!htbp]
\centering
\small
\begin{tabular}{lc}
\toprule
Configuration & BFCLv3 (en, multi-turn subset) \\
\midrule
\multicolumn{2}{l}{\textit{Step 1a: assistant text share (no irrelevance)}} \\
\textbf{20\%} & \textbf{23.38} \\
30\% & 19.37 \\
50\% & 15.75 \\
70\% & 11.62 \\
\bottomrule
\end{tabular}

\vspace{0.5em}

\begin{tabular}{lcc}
\toprule
Configuration & Irrelevance & Relevance \\
\midrule
\multicolumn{3}{l}{\textit{Step 1b: irrelevance share within 20\% text}} \\
5\% & 79.78 & 83.33 \\
\textbf{10\%} & \textbf{84.26} & \textbf{83.33} \\
20\% & 79.16 & 83.33 \\
30\% & 86.27 & 66.67 \\
\bottomrule
\end{tabular}
\caption{GRPO Step~1 mixture sweep on English-only data, run on the 8B SFT checkpoint. Step~1a varies the share of assistant text responses with no synthetic irrelevance injected, scored on the multi-turn subset of BFCLv3 (en). Step~1b fixes the text share at the Step~1a setting ($20\%$) and varies the irrelevance rate within the text responses, scored on the irrelevance and relevance subsets of BFCLv3 (en); the relevance subset is small ($N{=}18$ in BFCLv3), so its single-configuration differences should be read as trends, while the irrelevance subset is larger. Bold rows mark the configurations carried forward, not the per-column maxima.}
\label{tab:grpo_step1_ablations}
\end{table}

\paragraph{GRPO.}
GRPO uses a slice of each language pool reserved before the SFT mixture is drawn, disjoint from the SFT sample by construction, with the same length stratification. The reward is binary and follows Tool-N1~\citep{zhang2025tooln1}: for a tool-call target, the call is compared to ground truth as a multiset of \texttt{(name, arguments)} dictionaries, reward $1$ on exact match and $0$ otherwise; for a text target, reward $1$ if and only if no tool call can be parsed. Text quality is not otherwise rewarded --- the same opening the IF verifier left for empty completions --- but no degenerate-text collapse appeared here: the reward is indifferent among non-call outputs and, unlike the IF verifier, gives minimal completions no advantage, so the SFT text distribution is simply preserved; refusal and interpretive-reply quality is out of scope and covered by the in-house arena. This binary reward admits a dominant exploit: when in doubt, emit a call. After the first GRPO runs the model defaults to this low-risk action and irrelevance detection suffers; the 32B direction check quantifies the effect --- without injected irrelevance the subset sits at 78.3, versus 88.5 with it (Table~\ref{tab:grpo_32b_confirm}). We correct this through the data distribution rather than the reward, which we leave untouched --- more elaborate decompositions with separate format, syntax, and semantic components were explored in early iterations, but the extra terms added surface for reward hacking, so we retained plain exact match. We tune two knobs: the share of synthetic irrelevance cases, which counters over-calling and restores irrelevance detection, and the share of assistant text targets, which governs multi-turn accuracy. Irrelevance cases are constructed from existing samples by removing the ground-truth tool from the available pool and rewriting the target as a textual refusal --- the no-applicable-tool case of abstention.

\begin{table}[!htbp]
\centering
\small
\begin{tabular}{lcc}
\toprule
EN-RU split & BFCLv3 (en) & BFCLv3 (ru) \\
\midrule
40-60 & 60.22 & 55.44 \\
50-50 & 59.12 & 56.71 \\
60-40 & 59.44 & 57.22 \\
\textbf{70-30} & \textbf{62.62} & \textbf{57.05} \\
\bottomrule
\end{tabular}
\caption{GRPO Step~2 sweep of the English--Russian split on the 8B SFT checkpoint, with the text share fixed at $20\%$ and the irrelevance share at $10\%$ within text (the Step~1b setting from Table~\ref{tab:grpo_step1_ablations}). Bold marks the configuration carried forward, not the per-column maximum.}
\label{tab:grpo_step2_enru_ablation}
\end{table}

The mixture is tuned one axis at a time with the others fixed, first on English and then extended to Russian; throughout, we act on the sign and ordering of each axis rather than on point differences between adjacent configurations, which single runs per setting cannot support. Step~1a sweeps the assistant text share (Table~\ref{tab:grpo_step1_ablations}); multi-turn accuracy is highest at the smallest tested share and falls monotonically as the share grows. We adopt $20\%$ as a floor rather than a located optimum: shares below it were not searched, since the text targets are what supply abstention and interpretive-reply behaviour, and preserving them is a guardrail we impose rather than a trade-off we measured. Holding this fixed, Step~1b sweeps the synthetic-irrelevance fraction within the text targets; we take $10\%$, the interior peak of irrelevance detection across the $5$--$20\%$ range, and do not chase the higher value at $30\%$, where the small BFCLv3 relevance subset ($N{=}18$) loses three examples --- a drop we cannot distinguish from noise but see no reason to risk. Step~2 then fixes both and sweeps the English--Russian split (Table~\ref{tab:grpo_step2_enru_ablation}). Russian varies by at most $1.8$ points across the sweep --- if anything trending slightly higher at lower Russian shares, which we read as noise --- so the split is adopted as a default rather than a finding: 70-30, whose English advantage rests on a single configuration we do not over-read, keeps Russian within $0.2$ of its sweep peak. The resulting GRPO mixture is 70\% English and 30\% Russian; within each language, 80\% tool-call and 20\% text targets; within text targets, 10\% are synthesised irrelevance cases.

Because 8B signal does not always transfer to 32B, the sign of each operative axis is re-verified directly at 32B (Table~\ref{tab:grpo_32b_confirm}); the magnitudes are not re-optimised there, and the 8B mixture is carried forward as a default --- only the directions, which reproduce across scales and data pools, are load-bearing. Over-calling is an exploit of the reward structure, which is identical at both scales, so its direction carries even where the magnitude does not --- Table~\ref{tab:grpo_32b_confirm} reproduces both signs at 32B. The reward stays minimal; behaviour is shaped through the mixture.

\begin{table}[!htbp]
\centering
\small
\resizebox{\columnwidth}{!}{%
\begin{tabular}{lcccc}
\toprule
Model & BFCLv3 (en) & BFCLv3 (ru) & AceBench & $\tau^2$ \\
\midrule
Qwen3-32B (no-think)        & 63.13 & 54.03 & 54.60 & 31.53 \\
~~+ SFT                     & 69.37 & 58.69 & 67.80 & 35.20 \\
~~+ SFT + GRPO              & \textbf{73.19} & \textbf{66.93} & \textbf{73.00} & \textbf{38.20} \\
\bottomrule
\end{tabular}%
}
\caption{Per-stage contribution of the FC recipe in isolation: base Qwen3-32B (no-think), FC-only SFT, then GRPO from that SFT checkpoint. These isolate each stage's effect and are not the deployed model, which branches from the shared multi-domain SFT. Best values per column in bold.}
\label{tab:fc_results}
\end{table}
\paragraph{Findings.}
The per-stage breakdown shows where in the recipe the gains arise (Table~\ref{tab:fc_results}). On the English-side benchmarks (BFCLv3 en and AceBench) SFT accounts for the majority of the gain and GRPO adds a further increment, consistent with GRPO acting as a precision-tightening stage once the basic FC behaviour is in place. On Russian BFCLv3 the pattern inverts, with GRPO contributing its largest single-stage gain --- achieved despite the Russian share dropping from $50\%$ at SFT to $30\%$ at GRPO. We read this as the over-calling rebalancing rather than a language-share effect: the Step~2 sweep (at 8B) moves ruBFCLv3 by at most $1.8$ points as the Russian share varies from $60\%$ to $30\%$ (Table~\ref{tab:grpo_step2_enru_ablation}), so the share change cannot account for a gain of this size, whereas the rebalancing targets exactly the behaviours that move --- abstention (Table~\ref{tab:grpo_32b_confirm}) and multi-turn, with Russian multi-turn doubling over the recipe (Table~\ref{tab:fc_multiturn}). Gains on $\tau^2$ are more modest at every stage: it scores end-to-end success over a full session, whereas the per-turn reward optimises individual calls, not session outcomes; aligning the two is future work.

\begin{table}[!htbp]
\centering
\small
\begin{tabular}{lcc}
\toprule
Text share & Overall & Multi-turn \\
\midrule
30\%          & 68.9          & 38.0 \\
\textbf{20\%} & \textbf{70.7} & \textbf{44.8} \\
\bottomrule
\end{tabular}

\vspace{0.6em}

\begin{tabular}{lcc}
\toprule
Irrelevance share & Overall & Irrelevance \\
\midrule
0\%           & 71.4          & 78.3 \\
\textbf{10\%} & \textbf{72.7} & \textbf{88.5} \\
\bottomrule
\end{tabular}
\caption{Direction check of the two rebalancing axes at 32B on English BFCLv3 (best checkpoint by overall accuracy; the ordering is preserved under the mean over checkpoints), each axis swept independently on English data with the other settings fixed, so the \textit{Overall} columns are not directly comparable across the two sub-tables. A smaller text share yields higher multi-turn accuracy, and injecting synthetic irrelevance raises irrelevance detection. These are direction-confirmation runs on intermediate data pools (snapshots predating the final regeneration round), scored on BFCLv3 alone; they verify the sign of each axis and are not the per-stage results in Table~\ref{tab:fc_results}, which use the selected mixture and the full four-benchmark suite. Multi-turn here exceeds the final mixture's (Table~\ref{tab:fc_multiturn}), partly since the selected mixture trades some English multi-turn for Russian coverage and irrelevance handling, and partly since the pools differ.}
\label{tab:grpo_32b_confirm}
\end{table}

\section{Training Setup}
The SFT stage for the 32B model took 57 hours on 4 nodes with 8 H100 GPUs each. Training used gradient checkpointing, FSDP~\citep{zhao2023fsdp}, and sample packing into 32k-token contexts without truncation. The optimal fine-tuning hyperparameters are reported in Table~\ref{tab:sft_hparams}.

\begin{table}[!h]
\centering
\small
\setlength{\tabcolsep}{5pt}
\begin{tabular}{lc}
\toprule
\textbf{Hyperparameter} & \textbf{Value} \\
\midrule
Max sequence length & 32768 \\
Global train batch size & 32 \\
Device micro-batch size & 1 \\
Epochs & 2 \\
Precision & bf16 \\
Seed & 17 \\
\midrule
Optimizer & AdamW \\
Learning rate & $1 \times 10^{-6}$ \\
Adam $\beta_1, \beta_2$ & 0.9, 0.95 \\
Adam $\epsilon$ & $1 \times 10^{-12}$ \\
Weight decay & 0.0 \\
Scheduler & Cosine \\
Warmup ratio & 0.1 \\
Final LR multiplier & 0.1 \\
Gradient clipping & 2.0 \\
\bottomrule
\end{tabular}
\caption{Main SFT hyperparameters.}
\label{tab:sft_hparams}
\end{table}

GRPO training was performed using verl~\citep{verl_paper}. Each RL expert was trained on 4 nodes with 8 H100 GPUs per node. The IF, General, and Tool experts took approximately 28, 40, and 62 hours, respectively. The main expert-specific differences were the reward functions, KL coefficients, sequence lengths, and the number of optimization steps, while the shared GRPO setup is reported in Table~\ref{tab:rl_experts_hparams_shared}. The expert-specific hyperparameters are reported in Table~\ref{tab:rl_experts_hparams_specific}.

\begin{table}[!h]
\centering
\small
\resizebox{\columnwidth}{!}{%
\begin{tabular}{lccc}
\toprule
\textbf{Hyperparameter} & \textbf{General} & \textbf{IF} & \textbf{Tool-use} \\
\midrule
Max prompt length & 1024 & 1024 & 8192 \\
Max response length & 4096 & 4096 & 8192 \\
Max tokens / GPU & 12288 & 12288 & 16896 \\
KL loss coefficient & 0.01 & 0.001 & 0.001 \\
Total steps & 150 & 100 & 200 \\
Reward function & RM + length & Verifier + RM & Tool-call verifier \\
\bottomrule
\end{tabular}
}
\caption{Expert-specific GRPO hyperparameters.}
\label{tab:rl_experts_hparams_specific}
\end{table}

\begin{table}[!h]
\centering
\small
\setlength{\tabcolsep}{5pt}
\begin{tabular}{lc}
\toprule
\textbf{Hyperparameter} & \textbf{Value} \\
\midrule
Train batch size & 512 \\
Rollouts per prompt ($n$) & 8 \\
PPO mini-batch size & 64 \\
PPO micro-batch size / GPU & 1 \\
Learning rate & $1 \times 10^{-6}$ \\
Sampling temperature & 1.0 \\
Top-$p$ & 1.0 \\
PPO clipping range & [0.20, 0.28] \\
Entropy coefficient & 0 \\
Gradient clipping & 1.0 \\
Rollout backend & vLLM~\citep{kwon2023vllm} \\
Tensor parallel size & 4 \\
Distributed strategy & FSDP2 \\
Compute & 4$\times$8 GPUs \\
\bottomrule
\end{tabular}
\caption{Shared GRPO hyperparameters across all RL experts.}
\label{tab:rl_experts_hparams_shared}
\end{table}

\section{Merging}
\label{sec:app_merging}

\begin{table*}[!htbp]
\centering
\small
\begin{tabular}{lcccccccc}
\toprule
\multirow{2}{*}{Configuration}
& \multirow{2}{*}{ruIFEval}
& \multicolumn{2}{c}{BFCLv3}
& \multicolumn{2}{c}{Ru-arena-hard}
& \multicolumn{2}{c}{Inhouse arena} \\
\cmidrule(lr){3-4}
\cmidrule(lr){5-6}
\cmidrule(lr){7-8}
& & RU & EN & Score & Avg len & Score & Avg len \\
\midrule
\multicolumn{8}{l}{\textit{From mixed SFT}} \\
Baseline (mixed SFT)                           & 0.731 & 53.12 & 61.18 & 74.92 & 1388 & 54.81 & 308 \\
FC $+$ IF                                      & \textbf{0.801} & 49.00 & 54.51 & \textbf{84.04} & 1662 & 57.21 & 286 \\
FC $+$ IF $+$ general                          & 0.687 & \underline{54.08} & \textbf{62.32} & 82.72 & \textbf{1243} & \textbf{62.41} & \underline{295} \\
FC $+$ IF $+$ general, 1-dom/batch             & \underline{0.737} & \textbf{54.21} & \underline{62.10} & \underline{83.72} & \underline{1284} & \underline{61.1}  & 296 \\
\midrule
\multicolumn{8}{l}{\textit{From general GRPO}} \\
Baseline (general GRPO)                        & 0.727 & 52.45 & 60.82 & 77.80 & \underline{1266} & 58.31 & 270 \\
FC 80\% $+$ IF 20\%                            & \underline{0.780} & 53.94 & 63.71 & 83.14 & 1540 & 59.03 & 258 \\
FC 65\% $+$ IF 35\%                            & \textbf{0.800} & 55.16 & 63.81 & \underline{83.19} & 1630 & 61.57 & 306 \\
FC 50\% $+$ IF 50\%                            & 0.779 & 55.13 & 63.41 & 80.60 & 1419 & 60.50 & \underline{255} \\
FC 71\% $+$ IF 29\%\textsuperscript{$\dagger$}                          & \underline{0.780} & \underline{56.23} & \textbf{65.95} & \textbf{83.61} & 1634 & \underline{61.68} & 283 \\
\textbf{FC 71\% $+$ IF 29\%, with length penalty}\textsuperscript{$\dagger$} & 0.771 & \textbf{56.53} & \underline{64.26} & 80.23 & \textbf{1256} & \textbf{61.85} & \textbf{253} \\
\bottomrule
\end{tabular}
\vspace{2pt}

{\footnotesize $\dagger$ Trained with a $1.7\times$ larger training budget than the other FC/IF mixing-ratio configurations.}

\caption{Joint multi-domain GRPO sweep, run on Qwen3-8B for cost. The upper block starts from the mixed SFT checkpoint (FC$+$IF$+$general); the lower block starts from the general GRPO checkpoint trained on top of the same mixed SFT. \textit{1-dom/batch} denotes the schedule in which each batch contains samples from a single domain (no in-batch domain mixing); all other rows use in-batch mixing. \textit{Baseline} rows report the starting checkpoint of each block.}
\label{tab:joint_rl_sweep}
\end{table*}

The IF, FC, and general-domain experts come from three independent GRPO
stages atop a shared SFT checkpoint, each with a domain-specific reward.
Merging them is non-trivial: the three reward signals are structurally
different, and joint multi-domain GRPO consistently degraded IF
performance in preliminary experiments, as the model optimised the other
rewards while relaxing strict constraint satisfaction. Conversely,
prolonged single-domain RL caused noticeable regression on
general-purpose benchmarks, making any single expert impractical as the
production checkpoint directly. We therefore adopted a train-separately, merge-later strategy: forking from the shared
SFT checkpoint into three independent GRPO runs, one per domain, then
combining experts via SLERP of model
weights. This section describes the merging procedure and the ablations
used to settle on its configuration.

\paragraph{SLERP Merging}
We apply SLERP in two stages. First, the IF and FC experts are merged at
coefficient $t_1$, producing $\theta_{\mathrm{IF{+}FC}}$. This
intermediate checkpoint is then merged with the general expert at
coefficient $t_2$, yielding $\theta_{\mathrm{merged}}$. We picked this ordering empirically: with three experts every combination can
be tried, so we enumerated all of them and Table~\ref{tab:merge_order} reports
the alternatives. The layer-wise SLERP coefficients were likewise chosen by
grid search on validation splits, and we have no general rule for carrying
either choice over to a different set of experts.

\paragraph{Coefficient selection.}
Rather than a single scalar $t$ per stage, we set the SLERP coefficient
separately for different parameter groups within each layer. For the
first-stage IF\,+\,FC merge we use a layer-wise schedule with distinct
values for self-attention projections, MLP projections, and remaining
parameters (embeddings, norms). Attention coefficients are swept across
$\{0, 0.3, 0.5, 0.7, 1\}$ per block, MLP coefficients follow the
complementary pattern, and all other parameters use $t = 0.5$. The
second-stage merge with the general expert uses a uniform $t_2 = 0.8$.
All coefficients were chosen by grid search over the merged checkpoints.

\paragraph{Polishing stage.}
We experimented with a short SFT polish pass on
$\theta_{\mathrm{merged}}$ to recover surface-level formatting
consistency that drifted after interpolation (e.g., occasional
regressions in tool-call JSON formatting or structural markers in IF
responses). Several small general-domain data mixtures were tried; the
resulting checkpoints matched the unpolished merge on general metrics but
consistently showed minor IF and FC regressions. Because the polish
stage yielded no net improvement at extra compute cost, we excluded it
from the final pipeline.

\paragraph{Joint RL and expert merging.}
As an alternative to expert merging, we explored joint multi-domain GRPO
with IF and FC samples mixed in a single batch, each scored by its
domain-specific verifier and the IF reward rescaled to match the FC
range. Domain balance is thus controlled by the data mixing ratio rather
than explicit reward weighting. We swept the starting checkpoint, either
mixed SFT or general GRPO on top of it, the domain composition and
batching schedule, single-domain per batch or in-batch mixing, and the
FC/IF sampling ratio. Table~\ref{tab:joint_rl_sweep} reports the full
sweep. The strongest configuration was joint GRPO over the general GRPO
  checkpoint with a 71\%/29\% FC/IF ratio; as even this configuration
  required the warm start and a $1.7\times$ budget while remaining
  sensitive to the mixing ratio, we also replicated this sweep at the 32B scale.

\paragraph{Sequential and joint multi-SLERP.}
As an alternative to the two-stage sequential merge, we evaluated a
single-stage joint SLERP in which all three experts are combined
simultaneously with weights $w_\mathrm{IF}, w_\mathrm{FC},
w_\mathrm{gen}$ summing to one. The two variants are not equivalent,
since SLERP is non-linear in the parameter vectors, and they make
different implicit assumptions about how the three experts should be
balanced. Table~\ref{tab:merge_operator} compares the sequential variant
used in Our model against joint multi-SLERP and TIES-Merging~\citep{yadav2024ties} applied
jointly to all three experts; for each method we report the best
configuration found by the same grid search procedure described above.

\begin{table}[!htbp]
\centering
\small
\resizebox{\columnwidth}{!}{%
\begin{tabular}{lcccccc}
\toprule
\multirow{2}{*}{\textbf{Method}}
  & \multicolumn{2}{c}{\textbf{Arena}}
  & \multirow{2}{*}{\textbf{ruIFEval}}
  & \multicolumn{2}{c}{\textbf{BFCL}} \\
\cmidrule(lr){2-3} \cmidrule(lr){5-6}
  & Ru-Hard & In-House & & En & Ru \\
\midrule
Joint multi-SLERP            & 92.15          & 64.42          & 0.781          & 70.84 & 64.41 \\
Joint TIES-Merging         & 90.74          & 63.76          & 0.765          & 69.18 & 63.14 \\
Seq. SLERP (Ours) & \underline{93.87} & \textbf{69.57} & \textbf{0.799} & \textbf{72.27} & \textbf{65.96} \\
\quad + polish      & \textbf{94.12} & \underline{68.10} & \underline{0.795}          & \underline{71.83}          & \underline{65.40} \\
\bottomrule
\end{tabular}%
}
\caption{Comparison of merging strategies for combining three domain-specific experts (IF, FC, general). Each row reports the best configuration found via grid search over the method's hyperparameters. \textit{Seq.\ SLERP} is the two-stage sequential merge used in Our model; \textit{+ polish} adds a short SFT pass on top of it.}
\label{tab:merge_operator}
\end{table}

\paragraph{Effect of the polish stage.}
We evaluated several polish configurations, varying the data subset and
training duration. Polished checkpoints performed comparably on
general-domain benchmarks but showed consistent minor regressions on
IFEval and BFCLv3 relative to the unpolished merge.
Table~\ref{tab:merge_operator} summarises the best polish configuration
against the unpolished merge. Given no quality gain and additional
compute overhead, the polish stage was dropped from the final recipe.

\section{Additional Evaluations}
\label{sec:app_additional_evals}
\begin{table*}[!htbp]
\centering
\small
\begin{tabular}{lcccc}
\toprule
\multirow{2}{*}{\textbf{Model}} & \multirow{2}{*}{IFEval} & \multirow{2}{*}{MultiChallenge} & \multicolumn{2}{c}{Arena Hard 2} \\
\cmidrule(lr){4-5}
 &  &  & Hard & Creative \\
\midrule
Ours & \textbf{0.7872} & \underline{37.4} & 60.8 & \underline{70.4} \\
Qwen3-235B-A22B-Instruct-2507 & \underline{0.7798} & \textbf{43.6} & \textbf{71.0} & \textbf{87.4} \\
T-Pro-2.0 (think) & 0.7230 & 37.0 & \underline{63.1} & 58.6 \\
T-Pro-2.0 (no-think) & 0.7023 & 26.4 & 46.2 & 62.8 \\
Qwen3-32B (think) & 0.7785 & 30.8 & 45.2 & 50.6 \\
Qwen3-32B (no-think) & 0.7770 & 31.5 & 32.3 & 41.8 \\
\bottomrule
\end{tabular}
\caption{Comparison of models on English benchmarks.}
\label{tab:english}
\end{table*}

\paragraph{MultiChallenge} Tables~\ref{tab:dialogue-if-fc} and \ref{tab:english} report MultiChallenge average scores for six models in English and Russian. Qwen3-235B-A22B-Instruct-2507 leads in both languages with 43.6 and 46.2 respectively. Our model ranks second at 37.4 and 34.1, narrowly ahead of T-Pro-2.0 in thinking mode at 37.0 and 31.9. The two Qwen3-32B variants cluster around 30--31\% regardless of thinking mode, while T-Pro-2.0 without thinking trails at 26.4 and 27.8. The English-Russian gap varies across models, ranging from a 5-point drop for T-Pro-2.0 think to a 2.6-point gain for Qwen3-235B, indicating that cross-lingual robustness in multi-turn instruction following is not uniform across models.

\begin{table}[!htbp]
\centering
\small
\resizebox{\columnwidth}{!}{%
\begin{tabular}{lcccc}
\toprule
\multirow{2}{*}{\textbf{Model}} & \multicolumn{2}{c}{\textbf{Strict-IF}} & \multirow{2}{*}{\textbf{Loose-IF}} \\
\cmidrule(lr){2-3}
 & ex. & constr. & \\
\midrule
Ours & \textbf{0.81} & \textbf{0.89} & 0.58 \\
Qwen3-235B-A22B-Instruct-2507 & \textbf{0.82} & \underline{0.83} & \textbf{0.66} \\
T-Pro-2.0 (think) & 0.60 & 0.65 & \underline{0.59} \\
T-Pro-2.0 (no-think) & 0.51 & 0.50 & 0.48 \\
Qwen3-32B (think) & \underline{0.77} & 0.81 & 0.56 \\
Qwen3-32B (no-think) & 0.68 & 0.67 & 0.52 \\
\bottomrule
\end{tabular}%
}
\caption{In-house instruction-following benchmark results.}
\label{tab:inhouse-if-leaderboard}
\end{table}

\paragraph{Sensitivity to the judge model}
The in-house Arena is scored by an LLM judge, so we re-scored the cached
generations with three additional judges: GLM-4.6,
Mistral-Large-3-675B-Instruct-2512 \citep{mistrallarge3}, and DeepSeek-V3.1-Terminus \citep{deepseekai2024}
(Table~\ref{tab:judge-sensitivity}). We reused the same generations and ran
each judge deterministically, so the comparison isolates the judge itself.
Absolute scores move, but the ordering our claims rest on does not: the
deployed checkpoint ranks first under every judge, the same three models take
the top places in all four columns, and both of our checkpoints stay ahead of
Qwen3-32B in no-think mode. DeepSeek-V3, the judge behind the main results, is
not from the Qwen family our checkpoints derive from, so it cannot prefer them
out of family resemblance.

\begin{table*}[!htbp]
\centering
\small
\begin{tabular*}{\textwidth}{@{\extracolsep{\fill}}lcccc}
\toprule
\multirow{2}{*}{\textbf{Model}} & \multicolumn{4}{c}{\textbf{Judge}} \\
\cmidrule(lr){2-5}
 & DeepSeek-V3
 & GLM-4.6
 & Mistral-Large-3
 & DeepSeek-V3.1 \\
\midrule
T-Pro 2.1 (internal)          & \textbf{69.57}    & \textbf{68.18}    & \textbf{64.0}    & \textbf{58.8} \\
T-Pro 2.1 (public)            & \underline{66.80} & 66.24             & 60.4             & \underline{58.7} \\
Qwen3-235B-A22B-Instruct-2507 & 65.83             & \underline{67.85} & \underline{63.4} & 58.2 \\
T-Pro 2.0 (think)             & 61.17 & 62.20 & 58.5 & 52.3 \\
T-Pro 2.0 (no-think)          & 57.46 & 60.59 & 59.0 & 47.3 \\
Qwen3-32B (think)             & 60.46 & 63.08 & 57.8 & 54.8 \\
Qwen3-32B (no-think)          & 59.49 & 62.67 & 59.1 & 53.8 \\
\bottomrule
\end{tabular*}
\caption{In-house Arena scores under four judges, computed on identical
cached generations with deterministic judging. Best per column in bold,
second best underlined.}
\label{tab:judge-sensitivity}
\end{table*}

\paragraph{Inhouse IFEval} Post-training produces a sharp Strict-IF improvement and a smaller but positive Loose-IF gain (Table~\ref{tab:inhouse-if-leaderboard}). On Strict-IF, our model reaches 0.81 example-level and 0.89 constraint-level accuracy, against 0.68 and 0.67 for base no-think Qwen3-32B, matching the seven-times-larger Qwen3-235B-A22B-Instruct-2507 at the same parameter count. T-Pro 2.0 sits well below both base model modes, being thinking-optimised T-Pro 2.0 recipe traded base-skill quality for chain-of-thought capability, while the our recipe restores and surpasses both. On Loose-IF, our model reaches 0.58, ahead of base no-think Qwen3-32B at 0.52 but below thinking-mode T-Pro 2.0 at 0.59 by a margin within judge noise, and well below Qwen3-235B-A22B-Instruct-2507 at 0.66. The asymmetry is consistent with our reward design: the VerIF-style verifier reward is by construction concentrated on code-verifiable constraints, and gains transfer cleanly to Strict-IF on real production distribution. Loose constraints like tone, role, and semantic prohibitions are not directly optimised by the verifier reward, and the residual gap to a much larger generalist model indicates the natural next direction for the recipe.

\paragraph{Inhouse BFCL}
\begin{table}[!htbp]
\centering
\small
\resizebox{\columnwidth}{!}{%
\begin{tabular}{@{}lcccccc@{}}
\toprule
\textbf{Model} & \textbf{Ex.} & \textbf{Dec.} & \textbf{Tool} & \textbf{Req.} & \textbf{Sch.} & \textbf{Val.} \\
\midrule
Ours        & \textbf{0.79} & 0.88 & 0.73 & 0.73 & 0.98 & 0.80 \\
Qwen3-235B         & 0.77 & 0.87 & 0.70 & 0.70 & 0.99 & 0.83 \\
Qwen3-32B (no-think)                  & 0.71 & 0.79 & 0.61 & 0.61 & 0.99 & 0.81 \\
T-Pro 2.0 (no-think)                  & 0.68 & 0.75 & 0.55 & 0.55 & 0.98 & 0.79 \\
Qwen3-32B (think)                     & 0.67 & 0.76 & 0.54 & 0.54 & 0.97 & 0.86 \\
T-Pro 2.0 (think)                     & 0.67 & 0.77 & 0.53 & 0.53 & 1.00 & 0.85 \\
\bottomrule
\end{tabular}
}
\caption{In-house tool-calling benchmark. Ex. is the example-level pass rate; the remaining columns decompose it into call-vs-text decision (Dec.), tool(set) match (Tool), required-argument presence (Req.), schema validity (Sch.), and constrained-argument values (Val.). Free-text argument values and text-reply content are excluded from scoring by construction. Here Qwen3-235B  stands for Qwen3-235B-A22B-Instruct-2507.}
\label{tab:inhouse-fc-leaderboard}
\end{table}

Three observations from Table~\ref{tab:inhouse-fc-leaderboard} are
worth highlighting. Both modes of T-Pro 2.0 sit at the bottom of the
leaderboard, scoring 0.67 and 0.68, below Qwen3-32B in both think
and no-think configurations; T-Pro 2.0 was optimised primarily for
chain-of-thought reasoning, and the thinking-mode bias comes at a
measurable cost on single-step tool selection. Our updated recipe
recovers and surpasses the base: the final model in no-think mode
reaches an example pass of 0.79 at the same parameter count as
Qwen3-32B no-think at 0.71, matching the seven-times-larger
Qwen3-235B-A22B-Instruct-2507 at the top of the leaderboard, 0.79 to
0.77. The gain concentrates on the components our recipe targets
directly, call-or-text decision at +9 points over base and
tool/required-argument match at +12 points, while schema validity is
saturated for all candidates at $\geq 0.97$ and constrained-value
scores are flat across models. The component decomposition confirms
this pattern: schema validity is uniformly high, 0.97 to 1.00, and
uninformative, so the measured differences reflect the harder skills of
deciding whether to call, routing to the correct tool, and supplying
required arguments, which are exactly the skills the FC reward in
Section~\ref{sec:app_fc_expert} optimises.

\paragraph{SmartSearch}
Table ~\ref{tab:smartsearch_results_react} reports the SmartSearch results in a ReAct loop. Qwen3-235B-A22B-Instruct-2507 leads, reaching 0.551 recall, 0.852 grounded rate, and 0.669 F1$_{\mathrm{R\&G}}$. Our model ranks the second and improves over base no-think Qwen3-32B on every metric: recall from 0.350 to 0.434, grounded rate from 0.754 to 0.778, and F1$_{\mathrm{R\&G}}$ from 0.478 to 0.557. Despite running without reasoning, its F1$_{\mathrm{R\&G}}$ also exceeds both Qwen3-32B and T-Pro-2.0 in their thinking modes (0.491 and 0.537), indicating that on this task the function-calling recipe contributes more than test-time reasoning does for the baselines. With no task-specific training for this domain or tool setting, the result indicates that the general function-calling recipe transfers beyond static benchmarks to a realistic in-domain retrieval task.

\begin{table}[!htbp]
\centering
\small
\resizebox{\columnwidth}{!}{%
\begin{tabular}{lccc}
\toprule
\textbf{Model} & Recall & GR & F1$_{\mathrm{R\&G}}$ \\
\midrule
Ours              & \underline{0.434} & 0.778 & \underline{0.557} \\
Qwen3-235B-A22B-Instruct-2507      & \textbf{0.551} & \textbf{0.852} & \textbf{0.669} \\
T-Pro-2.0 (think)                  & 0.409 & \underline{0.780} & 0.537 \\
T-Pro-2.0 (no-think)               & 0.316 & 0.769 & 0.447 \\
Qwen3-32B (think)                  & 0.360 & 0.773 & 0.491 \\
Qwen3-32B (no-think)               & 0.350 & 0.754 & 0.478 \\
\bottomrule
\end{tabular}
}
\caption{SmartSearch results in a ReAct loop, where GR denotes Grounded Rate.}
\label{tab:smartsearch_results_react}
\end{table}

\end{document}